\documentclass[journal]{IEEEtran}
\usepackage{amsmath,amsfonts}
\usepackage{algorithmic}
\usepackage{algorithm}
\usepackage{array}
\usepackage[caption=false,font=normalsize,labelfont=sf,textfont=sf]{subfig}
\usepackage{textcomp}
\usepackage{stfloats}
\usepackage{url}
\usepackage{verbatim}
\usepackage{graphicx}
\usepackage{cite}
\usepackage{tabularx}
\usepackage{adjustbox}

\usepackage[super]{nth}
\usepackage{pifont}
\usepackage{color}
\usepackage{xcolor}
    \usepackage[]{nomencl}   
    \makenomenclature
    \usepackage{multirow}
    	\DeclareMathOperator*{\argmaxB}{argmax} 

	\usepackage{hyperref} 
\hypersetup{
	colorlinks=true,       
	linkcolor=red,        
	citecolor=blue,        
	filecolor=magenta,     
	urlcolor=black
}

\begin{document}

\title{ A Sparse Bayesian Learning for Diagnosis of Nonstationary and Spatially Correlated Faults with  Application to Multistation Assembly Systems}

\author{Jihoon Chung and Zhenyu (James) Kong
\thanks{The authors are with the  Department of Industrial and Systems Engineering, Virginia Tech, USA (email: jihoon7@vt.edu; zkong@vt.edu). (\textit{Corresponding author: Zhenyu James Kong.})
This work has been submitted to the IEEE for possible publication. Copyright may be transferred without notice, after which this version may no longer be accessible.}}

\markboth{}%
{Shell \MakeLowercase{\textit{et al.}}: A Sample Article Using IEEEtran.cls for IEEE Journals}


\maketitle
\vspace{-2cm}
\begin{abstract}
Sensor technology developments provide a basis for effective fault diagnosis in manufacturing systems. However, the limited number of sensors due to physical constraints or undue costs hinders the accurate diagnosis in the actual process. In addition, time-varying operational conditions that generate nonstationary process faults and the correlation information in the process require to consider for accurate fault diagnosis in the manufacturing systems. This article proposes a novel fault diagnosis method: clustering spatially correlated sparse Bayesian learning (CSSBL), and explicitly demonstrates its applicability in a multistation assembly system that is vulnerable to the above challenges.  Specifically, the method is based on a practical assumption that it will likely have a few process faults (sparse). In addition, the hierarchical structure of CSSBL has several parameterized prior distributions to address the above challenges. As posterior distributions of process faults do not have closed form, this paper derives approximate posterior distributions through Variational Bayes inference. The proposed method's efficacy is provided through numerical and real-world case studies utilizing an actual autobody assembly system. The generalizability of the proposed method allows the technique to be applied in fault diagnosis in other domains, including communication and healthcare systems.

\end{abstract}

\def\abstractname{Note to Practitioners}
\begin{abstract}
This article proposes a new process fault diagnosis method: clustering spatially correlated sparse Bayesian learning.
This method effectively diagnoses time-varying defects by leveraging the correlation structures in the process when sensor measurements are insufficient. The actual autobody assembly process is utilized to show the proposed method's effectiveness. The proposed method performs superior to the benchmark methods in fault detection capability. In addition, the proposed method accurately estimates the severity of the process faults, providing significant information to the practitioners for their decision-making in the maintenance schedule. Specifically, the error between the estimation from the proposed method and the actual severity of the process faults achieves less than 10\% of error of all the benchmark methods when there exists a high correlation between the variations of the fixture locators in the autobody assembly system.

\end{abstract}


\section{Introduction}\label{s:sec.1}

\IEEEPARstart{T}he sparse estimation has received considerable attention in signal processing due to its ability to reconstruct a high-dimensional sparse source signal from a low-dimensional measurement \cite{shirazinia2013analysis}. Specifically,  sparse estimation has broad applications in a wide range of industries in identifying the sources of sensor measurements. These applications include channel estimation in wireless systems \cite{dai2018joint}, Electroencephalography (EEG) source localization in neuroimaging \cite{gorodnitsky1995neuromagnetic}, radar detection \cite{ender2010compressive},  and fault diagnosis in manufacturing systems \cite{bastani2012fault}. \textcolor{black}{However, time-varying operational conditions in the manufacturing systems cause nonstationary process faults, hindering the accurate sparse estimation. For example, the component degradation (e.g., fixture wearing in assembly systems) will vary over time, violating the stationary process faults assumptions of most existing sparse estimation methods \cite{li2019condition, bastani2016compressive}.} In addition,  the correlation information that occurred due to the structure of the manufacturing system should consider for effective fault diagnosis \cite{bastani2018fault}, but it is often neglected in sparse estimation \cite{chung2023novel, lee2020variation}.

One motivating example to address the above challenges is the process fault diagnosis in the multistation assembly systems. The systems perform assembly operations from multiple stations to assemble a final product. The final product's quality relies on several factors known as key control characteristics (KCCs) \cite{bastani2018fault}. The positioning accuracy of fixture locators is KCCs in the multistation assembly \cite{bastani2012fault}. Fixture locators carry out the clamping of parts during the assembly process. Therefore, any deviations from their nominal positions can lead to dimensional quality issues in the final product. 
Hence, fault diagnosis in \textcolor{black}{multistation} assemblies estimates the mean and variance of KCCs, namely, the variations of fixture locators \cite{chung2023novel}. This article focuses on process faults due to excessive variance, which is a more challenging task to diagnose than mean shifts \cite{bastani2016compressive}.

Since monitoring the dimensional variation of KCCs is not feasible due to physical constraints in the process \cite{lee2020variation},  the key product characteristics (KPCs), which are essentially measurements obtained from the final product, can be utilized to estimate the variance of  KCCs. Specifically,  a fault-quality linear model represents the relationship between KPCs and KCCs as follows \cite{huang2007stream,huang2007stream1}:
\begin{equation}\label{eq:1}
    \textbf{y}=\Phi \textbf{x} +\textbf{v},
\end{equation}
where $\textbf{y}\in \mathbb{R}^{M \times 1} $represents $M$ dimensional measurements (i.e., KPCs), $ \textbf{x}\in \mathbb{R}^{N \times 1} $denotes $N$ KCCs, $\Phi \in \mathbb{R}^{M\times N}$ is a fault pattern matrix. The matrix contains all the process information of the multistation process, and $\textbf{v} \in \mathbb{R}^{M \times 1}$ denotes the noise. 
Process fault refers to KCCs (elements of $\textbf{x}$ in Eq.~\eqref{eq:1}) whose variance exceeds the design specifications.

\textcolor{black}{Numerous studies have employed Eq.~\eqref{eq:1} on fault diagnosis in multistation assembly systems\cite{ding2002fault,kong2008multiple}. 
 However, they assume that the number of measurements ($M$) is greater than the number of KCCs ($N$) which may not hold in actual manufacturing applications. This is because using an excessive number of sensors (measurements) can result in undue costs \cite{bastani2018fault}. If this assumption is violated, Eq.~\eqref{eq:1} becomes an underdetermined system that has a non-unique solution. To address this challenge, the sparse solution assumption \cite{candes2009near} that \textbf{x} in Eq.~\eqref{eq:1} has a minimal number of non-zero elements is required. In the fault diagnosis problem, sparsity refers to the small number of process faults in the fault-quality linear model, which is a reasonable assumption since there are typically only a few process faults in practice \cite{bastani2012fault}. }
Among the several sparse estimation methods, the Bayesian approach called sparse Bayesian learning has received much attention recently because of its superior estimation performance guaranteed from the several theoretical properties \cite{zhang2011sparse, tibshirani1996regression,candes2008introduction}.

Sparse Bayesian learning has been applied for fault diagnosis in \textcolor{black}{multistation assembly systems \cite{lee2020variation,li2016bayesian}}. These studies successfully identified process faults by providing prior distribution of KCCs (i.e., \textbf{x} in Eq.~\eqref{eq:1}) to encourage the sparsity of process faults. Especially the work in \cite{lee2020variation} applied Bayesian learning to diagnose the process faults based on the following multiple measurements vectors (MMV) model in sparse Bayesian learning \cite{cotter2005sparse}:
\begin{equation}\label{eq:3}
      \textbf{Y}=\Phi \textbf{X} +\textbf{V}.
\end{equation}
$\textbf{Y}=[\textbf{y}_{1},...,\textbf{y}_{L}] \in \mathbb{R}^{M\times L}$ is a measurement matrix consisting of \emph{L} KPCs samples collected over $L$ time periods, where $\textbf{y}_{k}$ denotes $k^{th}$ KPCs sample. $\textbf{X}=[\textbf{x}_{1},...,\textbf{x}_{L}] \in \mathbb{R}^{N\times L}$ is a matrix, where $\textbf{x}_{k}$ is a vector that represents KCCs of $k^{th}$ KPCs sample. $\textbf{V}\in\mathbb{R}^{M\times L}$ is a noise matrix.  \textcolor{black}{\cite{lee2020variation} effectively identified the process faults based on the assumption that $L$ KPCs samples share the same process faults. In other words, the process has stationary process faults in $L$ time periods.} However, \cite{lee2020variation} did not consider the spatial correlation between KCCs (i.e., fixture locators). For example, the fixture locators physically located on the same part would vary together \cite{bastani2018fault}.
\begin{figure}[!htp]
    \centering
    \includegraphics[width=0.5\textwidth]{{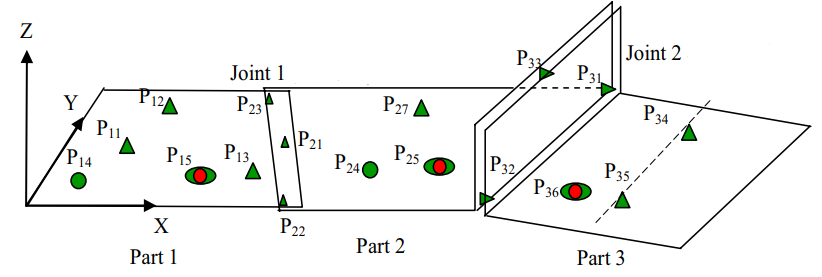}}\vspace{-0.0cm}
    \caption{A three-part assembly process is illustrated. $P_{ij}$ shows the $i^{th}$ fixture locator on the $j^{th}$ part \cite{kong2006mode}.}
    \label{fig:fig1}
\end{figure}
This phenomenon is depicted in Fig.~\ref{fig:fig1} for multistation assembly systems, where $P_{ij}$ corresponds to the $i^{th}$ fixture locator on the $j^{th}$ part.
Consequently, any deviations in the locator $P_{22}$ can result in variations to the nearby locator $P_{23}$ since they are situated nearby and share the same parts \cite{kong2006mode, bastani2018fault}.

Beyond the spatial correlation between KCCs, the nonstationarity of process faults also needs to be considered for accurate fault diagnosis. In practice, the external time-varying operational condition, including temperature and noise, may cause nonstationary process faults over time in the manufacturing systems \cite{li2019condition}. \textcolor{black}{For example, the fixture locators in the assembly process can be deviated because of the thermal expansions of pins caused by the ambient temperature of the process that varies over time  \cite{abellan2012state}.} In addition, the variation of some locators will be propagated to other locators over time if the process faults are not mitigated immediately \cite{shi2022process}.
\textcolor{black}{It results in nonstationary process faults along the KPCs samples collected over time. In other words, the KCCs with excessive variance differ depending on the columns of $\textbf{X}$ in Eq.~\eqref{eq:3}. Therefore, the stationary process faults assumption applied in the previous studies  \cite{lee2020variation, bastani2018fault} needs to be addressed.}

To consider both the spatial correlation of KCCs and the nonstationary process faults along the KPCs samples, this paper proposes a novel sparse Bayesian learning method, namely, clustering and spatially correlated sparse Bayesian learning (CSSBL).
Given KPCs samples, the proposed method   clusters the samples into groups sharing the same process faults. At the same time, our method estimates the variance of KCCs of each group to identify the process faults accurately. The following summarizes the contributions of this study: 

\vspace{-0.0cm}\begin{itemize}
\item From the methodological point of view, this paper proposes a novel sparse Bayesian learning that can consider both the spatial correlation of KCCs and the nonstationary process faults. Since the posterior distribution of the sparse solution in the proposed method is computationally intractable, this article derives the approximate posterior distribution of the sparse solution via Variational Bayes inference \cite{petersen2005slow}.\vspace{-0.0cm}
\item From the application perspective, the proposed method is applied to fault diagnosis in the actual auto-body assembly process.  The method performs better for process fault detection capability than the benchmark methods. Furthermore, the proposed method accurately estimates locators' variance representing the severity of the process faults to assist practitioners' decision-making in their maintenance policy.
\end{itemize}

\vspace{-0.0cm}The subsequent sections of this paper are structured in the following manner. Section~\ref{s:sec2} presents an overview of relevant literature, while Section~\ref{s:sec3} introduces the proposed methodology. The methodology's effectiveness is evaluated through numerical case studies in Section~\ref{s:sec4}. Section~\ref{s:sec5} provides real-world case studies which are fault diagnosis problems in the multistation assembly system. Finally, Section~\ref{s:sec6} discusses conclusions and future work.

\section{Review of Related Work} \label{s:sec2}
The related existing studies of fault diagnosis in manufacturing systems are reviewed in Section~\ref{s:sec2.1}. Then, the literature related to sparse Bayesian learning is provided in Section~\ref{s:sec2.2}. Afterward, the research gaps in the current work are identified in Section~\ref{s:sec2.3}. 
\subsection{\emph{Fault Diagnosis Methodologies in Manufacturing Systems}} \label{s:sec2.1}
Numerous studies have focused on fault diagnosis methodologies for manufacturing systems, utilizing the fault-quality model outlined in Eq.~\eqref{eq:1}. \cite{kong2008multiple} developed a PCA-based orthogonal diagonalization strategy to transform the measurement data. It enabled the estimation of the variance of KCCs in a multistation assembly system. \cite{ding2002fault} presented a fault diagnosis method in the multistation assembly systems integrating the state space model of the process and matrix perturbation theory. \cite{wang2005multi, wang2006error} proposed a fault diagnosis method in the machining process considering the process physics regarding how fixtures generate the patterns. Using this method, root cause identification was conducted sequentially. The approaches mentioned above assume that the number of measurements is greater than the number of KCCs (i.e., $M>N$ in Eq.~\eqref{eq:1}). However, this assumption may not always be consistent with industrial practice. These approaches become ineffective when this assumption is violated because the fault-quality linear model leads to an underdetermined system, resulting in a non-unique solution.

\textcolor{black}{To overcome an underdetermined system in the fault-quality linear model, sparse learning can be utilized, which has gained considerable attention in fault diagnosis and detection within manufacturing systems.} For fault diagnosis in the manufacturing system, \cite{han2018intelligent} developed a fault diagnosis method using dictionary learning and sparse representation-based classification. \cite{wang2021holistic} proposed a novel root cause diagnostic framework satisfying the assumption that sparse inputs affect the process output.  \cite{dai2021group} proposed a group-sparsity learning approach for bearing fault diagnosis. \textcolor{black}{In addition to fault diagnosis and detection in the general manufacturing system, sparse learning has been widely utilized to address the issue of an underdetermined system in the multistation assembly system.} In particular, sparse Bayesian learning has been widely utilized to incorporate the sparsity of process faults as the prior distribution. \cite{bastani2012fault} proposed a fault diagnosis approach by enhancing the relevant vector machine to detect process faults using the sparse estimate of the variance change of KCCs. \cite{bastani2018fault} developed a spatially correlated sparse Bayesian learning to consider the spatial correlation of KCCs in sparse estimation. The work identifies the KCCs with mean shifts.  \cite{lee2020variation} proposed a Bayesian model to identify the KCCs that have variance increases in a multistation manufacturing system using the sparse variance component prior. \cite{chung2023novel} developed a novel sparse Bayesian learning to figure out the KCCs with mean shifts by considering the temporal correlation of KCCs and the prior knowledge of process faults.
\vspace{-0.3cm}
\subsection{\emph{Sparse Bayesian Learning }} \label{s:sec2.2}
After the introduction of sparse Bayesian learning (SBL) by \cite{tipping2001sparse}, many researchers have extended this approach significantly. For instance, \cite{wipf2009solving} was the first to apply SBL to sparse estimation for the single measurement vector model given in Eq.~\eqref{eq:1}. Subsequently, \cite{wipf2007empirical} further extended it to the MMV model (Eq.~\eqref{eq:3}) by developing the MSBL algorithm under the common support assumption. The notable advantage of SBL and MSBL is that their global minimum is always the sparsest solution, while that of the minimization-based sparse algorithms \cite{tibshirani1996regression}  \cite{chen2001atomic} is usually not the sparsest solution \cite{zhang2011sparse, candes2008enhancing}.
Based on the MMV model, many previous studies exploit the spatial correlation in solution vectors (i.e., rows in  \textbf{X} in Eq.~\eqref{eq:3}). \cite{zhang2013extension} proposed a block structure to exploit the intra-block correlation for sparse estimation. \cite{fang2014pattern} developed a Bayesian method for recovery of block-sparse solution whose block-sparse structures are entirely unknown. \cite{han2018bayesian} modeled the spatial structure of the solution as Markov dependency by the Beta process. Besides considering the spatial correlation of sparse solutions, work that considers nonstationary sparse solutions has been studied recently under the MMV model in the SBL framework. \cite{wang2015novel} developed a method using the Dirichlet process to cluster the measurements into groups with common sparsity patterns. \textcolor{black}{Compared to \cite{wang2015novel}, \cite{dai2018joint} proposed a more general method having two sparsity components: a commonly shared sparsity and an individual sparsity to deal with outliers that deviated from the uniform sparsity pattern in each group.} 
\vspace{-0.3cm}\subsection{
\emph{Research Gap Analysis }} \label{s:sec2.3}
\textcolor{black}{The research presented in Section~\ref{s:sec2.1} focuses on utilizing the sparsity of process faults for accurate fault diagnosis when low dimensional measurements exist in actual manufacturing systems.} However, there is a lack of efforts to identify process faults by considering the spatial correlation and the nonstationary process faults, which is common in industrial practice. Section~\ref{s:sec2.2} introduces methods in SBL. It introduces the work concerning the spatial correlation in the solution vector and dealing with the nonstationary sparse solution individually. However, it still lacks the work that uses both properties simultaneously. Therefore, this paper proposes a novel SBL method considering both properties for accurate fault diagnosis.

\vspace{-0.3cm}
\section{Proposed Research Methodology} \label{s:sec3}
This section proposes a novel sparse Bayesian hierarchical method: clustering and spatially correlated sparse Bayesian learning (CSSBL). The proposed CSSBL is described in Section~\ref{s:methods.3.1}, followed by Bayesian inference in Section~\ref{s:methods.3.2}.
\vspace{-0.6cm}
\subsection{\emph{Proposed Methodology}} \label{s:methods.3.1}
The proposed methodology is a sparse Bayesian hierarchical model considering the spatial correlations of KCCs and nonstationarity of process faults \textcolor{black}{along the KPCs samples. From Eq. \eqref{eq:3}, fault quality linear model of $k^{th}$ KPCs sample ($\textbf{y}_{k}$) can be written as follows:}
\begin{equation}\label{eq:4}
    \textbf{y}_{k}=\Phi \textbf{x}_{k}+\textbf{v}_{k},
\end{equation}
where $\textbf{x}_{k}$ is KCCs from $k^{th}$ KPCs sample. $\textbf{v}_{k}$ indicate the noise of $k^{th}$ KPCs sample following the Gaussian distribution with precision variable $\alpha$ \cite{dai2018joint, chung2023novel}. Therefore, the Gaussian likelihood is provided to $\textbf{y}_{k}$ in Eq.~\eqref{eq:4} as follows: 
\begin{equation}\label{eq:5}
    p(\textbf{y}_{k}|\textbf{x}_{k};\alpha) \sim N(\Phi \textbf{x}_{k}, \alpha^{-1}\textbf{I}_{M}).
\end{equation}

\begin{figure}[h]
    \centering
{\includegraphics[width=0.5\textwidth]{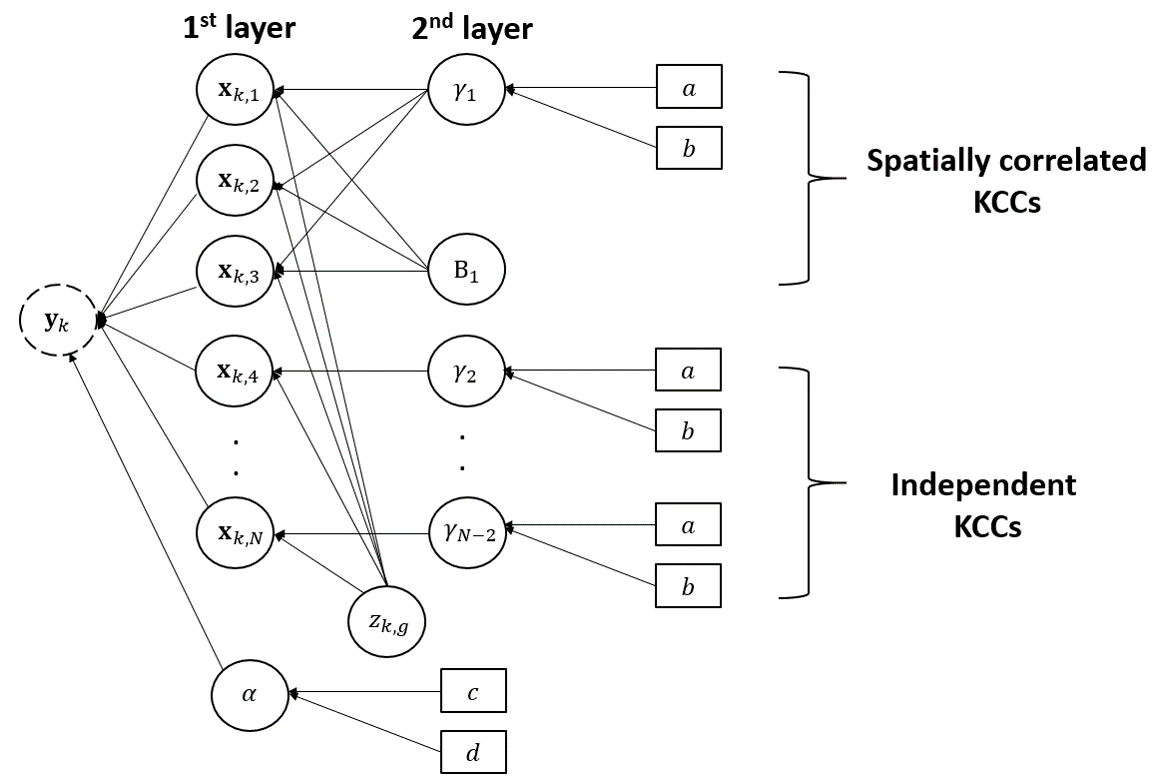} }\vspace{-0.4cm}
    \caption{Graphical representation of the proposed method, where  $\textbf{x}_{k}=\textbf{x}_{k,1},\textbf{x}_{k,2},...,\textbf{x}_{k,N}]$. A circle indicates a random variable or hyperparameter that needs to be estimated. A dashed circle and a square represent an observation and a constant, respectively.}
    \label{fig:fig3}\vspace{-0.2cm}
\end{figure}

The prior distributions in the proposed method consist of the following two hierarchical layers. \textcolor{black}{The first layer provides several prior distributions. First, the prior distribution of precision variable $\alpha$ in Eq.~\eqref{eq:5} is provided. Second, the prior distribution exploiting the spatial correlation of KCCs is offered. Finally, the prior distribution in the first layer clusters the KPCS samples into groups sharing the same process faults.} The second layer consists of prior distribution encouraging the sparsity of process faults. Fig.~\ref{fig:fig3} describes a graphical representation of the hierarchical layers in the proposed method. 

In the first layer of the hierarchical model in Fig.~\ref{fig:fig3},  the Gamma distribution is provided as the prior distribution of $\alpha$ in Eq.~\eqref{eq:5} as follows:
\begin{equation*}
    p(\alpha)=\Gamma(\alpha|c,d),
\end{equation*}
where $c$ and $d$ are some small constants to provide non-informative prior distribution (e.g., $c=d=10^{-4}$)) \cite{dai2018joint}. 
  Since the Gaussian distribution is widely used as the prior distribution for KCCs in the literature and practice \cite{lee2020variation, bastani2012fault}, the Gaussian distribution  is used for the prior distribution of KCCs from $k^{th}$ KPCs sample, $\textbf{x}_{k}$, as shown in Eq.~\eqref{equation:666}.
  
\vspace{-0.0cm}\begin{equation}\label{equation:666}
p(\textbf{x}_{k})=N(\textbf{x}_{k}|\boldsymbol{0},\text{C}_{k}).\vspace{-0.0cm}\
\end{equation}

\begin{figure}[h]
    \centering
    {\includegraphics[width=0.5\textwidth]{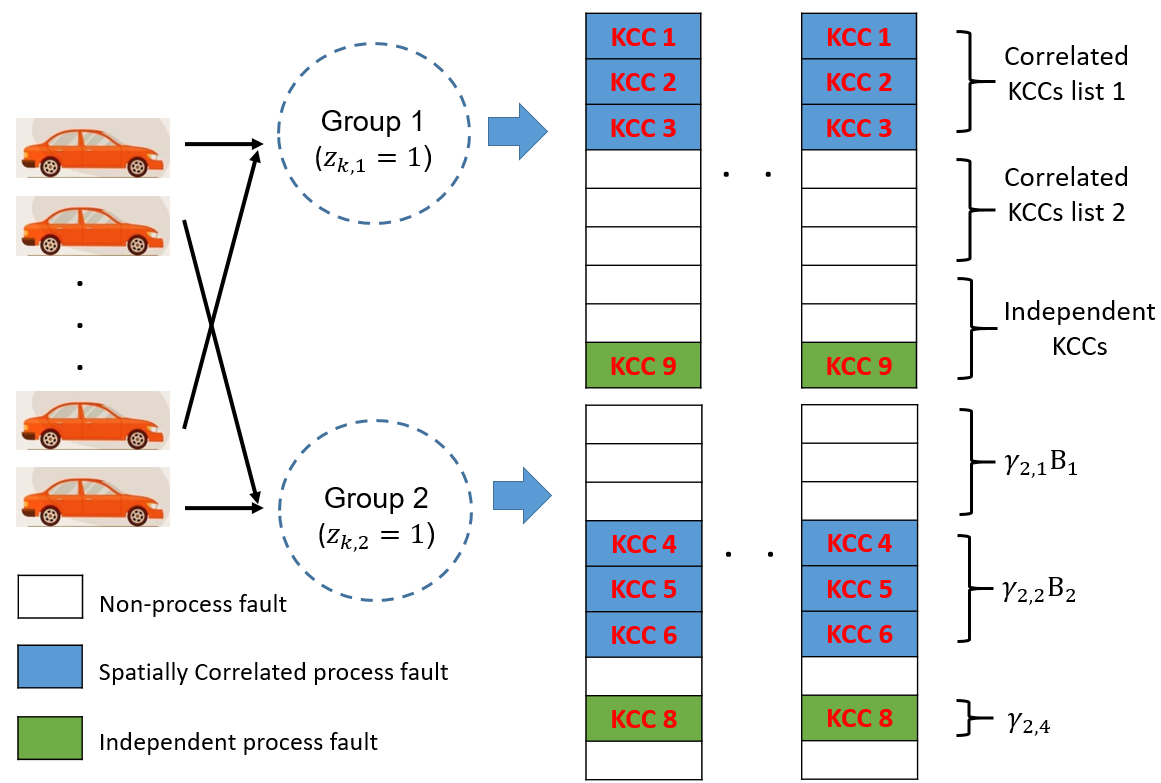}}\vspace{-0.0cm}
    \caption{Description of the proposed method when two correlated lists, each of which has a size of three KCCs, and two groups sharing the same process faults exist.}
    \label{fig:fig4}
\end{figure}

\textcolor{black}{To handle the spatial correlation of KCCs, the proposed approach considers two types of correlation structures of fixture locators in multistation assembly systems.
One is an independent locator whose deviation is uncorrelated to the deviation of other locators (i.e., independent tolerance mode in \cite{kong2006mode}). The other is correlated locators (i.e., composite tolerance mode in \cite{kong2006mode}). Specifically, the locators in the correlated list vary with a certain correlation with other locators in the list \cite{kong2006mode}.}
To consider this structure, the proposed method models the covariance matrix ($\text{C}_{k}$) of the prior distribution of $\textbf{x}_{k}$ in Eq.~\eqref{equation:666} as follows:
\begin{equation}\label{eq:6}
  \text{C}_{k}=
  \left[\begin{smallmatrix}
   (\gamma_{1}\text{B}_{1})^{-1}  & &  & & & \\
   &  \ddots &    & & & \\
   & & (\gamma_{r}\text{B}_{r})^{-1} & & & \\
   &  &  & (\gamma_{r+1}\text{B}_{r+1})^{-1} & & \\
   &  &  & & \ddots & \\
   &  &  & & &  (\gamma_{\text{R}}\text{B}_{\text{R}})^{-1} 
  \end{smallmatrix}\right].
\end{equation}
Matrix $\text{B}_{i}^{-1}$  $(\forall i =1,...,r)$ represents the correlation matrix of $i^{th}$ list of correlated KCCs with the size of $d_{i}$. The correlation structure of the $i^{th}$ list of correlated KCCs in  the multistation assembly system is shown as follows \cite{kong2006mode, bastani2018fault}:
\vspace{-0.0cm}\begin{equation}\label{eq:7}
  \text{B}_{i}^{-1}=
  \begin{bmatrix}
   1  & k_{i} & \cdots & k_{i} \\
   k_{i}  & 1& \cdots & \vdots \\
   \vdots  & k_{i} & \ddots & k_{i} \\
   k_{i}  & \cdots & k_{i} & 1 
  \end{bmatrix}_{d_{i}\times d_{i}} (\forall i=1,...,r),\vspace{-0.0cm}
\end{equation}
where $k_{i}$ denotes the correlation coefficient for the $i^{th}$ $(\forall i =1,...,r)$  correlated list of KCCs. In contrast, $\text{B}_{i}^{-1}$  $(\forall i=r+1,...,\text{R})$ represents the correlation matrix for the independent KCCs. Hence, $\text{B}_{i}^{-1}=\textbf{I}_ {1\times1} (\forall i =r+1,...,\text{R})$ and   $\Sigma_{i=1}^{\text{R}}d_{i}=N.$ $\gamma_{i}$ denotes the precision of the prior distribution of the $i^{th}$ list of KCCs. Fig.~\ref{fig:fig4} illustrates the proposed method when two correlated lists with the size of three KCCs exist, respectively (i.e., $r=2$ and $d_{1}=d_{2}=3$) and three independent KCCs (i.e., $\text{R}=5$).
Therefore, the prior distribution of KCCs from the $k^{th}$ KPCs sample considering the spatial correlation is defined as follows:
\vspace{-0.0cm}\begin{equation}\label{equation:6}
p(\textbf{x}_{k}|\Bar{\boldsymbol{\gamma}},\textbf{{B}})=N(\textbf{x}_{k}|\boldsymbol{0},\text{C}_{k}),\vspace{-0.0cm}\
\end{equation}
where $\Bar{\boldsymbol{\gamma}}=[\gamma_{1}, \gamma_{2},..., \gamma_{\text{R}}]$, and $\textbf{B}=[\text{B}_{1}, \text{B}_{2},...,\text{B}_{\text{R}}]$.

To consider the nonstationarity of process faults, the proposed method clusters KPCs samples into $G$ groups that share the same process faults. To achieve this objective, $\textbf{z}_{k}=[z_{k,1}, z_{k,2},..., z_{k,G}]$, the assignment vector for the $k^{th}$ KPCs sample is introduced. Specifically, if the $k^{th}$ KPCs sample belongs to the $g^{th}$ group, $\textbf{z}_k$ is a zero vector except for the $g^{th}$ element ($z_{k,g}$) being one. \textcolor{black}{For example, Fig.~\ref{fig:fig4} shows the case when two groups exist ($G=2$). If the $k^{th}$ sample belongs to Group 2, $\textbf{z}_k$ is a zero vector except for the $z_{k,2}$ being one.} Based on Eq.~\eqref{equation:6}, the prior distribution of $\textbf{x}_{k}$ in the proposed method is provided as follows by including the group index to covariance matrix ($C_{k}$ in Eq.~\eqref{equation:6}) and  assignment vector ($\textbf{z}_k$).
\vspace{-0.00cm}\begin{equation}\label{equation:7}
p(\textbf{x}_{k}|\textbf{z}_{k},\textbf{C}_{k})=\prod_{g=1}^{G} [N(\textbf{x}_{k}|\boldsymbol{0}, \text{C}_{k,g})]^{z_{k,g}},\vspace{-0.0cm}
\end{equation}
\textcolor{black}{where $\textbf{C}_{k}=[\text{C}_{k,1}, \text{C}_{k,2},..., \text{C}_{k,G}]$. \textcolor{black}{$\text{C}_{k,g}$ denotes the covariance matrix of the prior distribution for KCCs of the $k^{th}$ KPCs sample belonging to the $g^{th}$ group.} $\text{C}_{k,g}$ is defined  as follows: }
\begin{equation*}
 \text{C}_{k,g}=
  \left[\begin{smallmatrix}
   (\gamma_{g,1}\text{B}_{1})^{-1}  & &  & & & \\
   &  \ddots &    & & & \\
   & & (\gamma_{g,r}\text{B}_{r})^{-1} & & & \\
   &  &  & (\gamma_{g,r+1}\text{B}_{r+1})^{-1} & & \\
   &  &  & & \ddots & \\
   &  &  & & &  (\gamma_{g,\text{R}}\text{B}_{\text{R}})^{-1}  
  \end{smallmatrix}\right].
\end{equation*}
\textcolor{black}{Since each group has different process faults (KCCs with excessive variances), the variances of KCCs need to be indexed by groups. Therefore, $\Bar{\boldsymbol{\gamma}}=[\gamma_{1}, \gamma_{2},..., \gamma_{R}]$ in $C_{k}$ in Eq.~\eqref{equation:6} is replaced with $\bar{\boldsymbol{\gamma}}_{g}=[\gamma_{g,1}, \gamma_{g,2},..., \gamma_{g,R}]$, in $C_{k,g}$ in Eq.~\eqref{equation:7}.}

In the second hierarchical layer of the proposed method in Fig.~\ref{fig:fig3}, the Gamma distribution is provided as the prior distribution of $\bar{\boldsymbol{\gamma}_{g}}$ as follows to provide the sparsity of KCCs:
\begin{equation}\label{equation:8}
    p(\Bar{\boldsymbol{\gamma_{g}}})=\Pi_{r=1}^{\text{R}}\Gamma(\gamma_{g,r}|a,b),
\end{equation}
where $a$ and $b$ are some small constants (e.g., $a=b=10^{-4}$) \cite{dai2018joint}. The two-stage hierarchical structure consisting of Eqs.~\eqref{equation:7} and~\eqref{equation:8} is widely used to encourage the sparsity of  $\textbf{x}_{k}$ since the structure provides the prior distribution of $\textbf{x}_{k}$ that has a sharp peak at zero \cite{wipf2004sparse, ji2008bayesian}. In addition, the Gamma distribution in Eq.~\eqref{equation:8} enables the tractable inference of the approximate posterior distribution of $\bar{\boldsymbol{\gamma_{g}}}$ in the following section.

\vspace{-0.6cm}
\subsection{\emph{Bayesian Inference of the Proposed Methodology}} \label{s:methods.3.2}
The proposed method in Section~\ref{s:methods.3.1} has several hidden variables that need to be estimated from the KPCs samples ($\textbf{Y}=\{\textbf{y}_{k}\}_{k=1}^{K}$). Specifically, the KCCs $(\textbf{X}=\{\textbf{x}_{k}\}_{k=1}^{K})$, the variance of KCCs $(\Gamma = \{ \boldsymbol{\bar{\gamma_{g}}} \}_{g=1}^{G})$, the variables to \textcolor{black}{cluster the KPCs samples} ($\textbf{Z}=\{\textbf{z}_{k}\}_{k=1}^{K}$), and measurement noise in KPCs sample ($\alpha$) need to be estimated. In addition, the hyperparameters related to spatial correlations of KCCs ($\textbf{B}=\{\text{B}_{i}\}_{i=1}^{\text{R}}$) also need to be estimated. 

\textcolor{black}{Due to the complexity of the proposed hierarchical model in Section~\ref{s:methods.3.1}, the posterior distribution of hidden variables in the proposed method (Eq.~\eqref{eq:11}) does not have a closed form. Specifically, the denominator in Eq.~\eqref{eq:11} cannot be derived as a closed form.
\begin{align}\label{eq:11}
         P(\textbf{X}, \textbf{Z}, \Gamma, \alpha | \textbf{Y}) & = \frac{P(\textbf{Y}, \textbf{X}, \textbf{Z}, \Gamma, \alpha)}{  \idotsint\limits_{\textbf{X}, \textbf{Z}, \Gamma, \alpha } P(\textbf{Y}, \textbf{X}, \textbf{Z}, \Gamma, \alpha) \, d\textbf{X} \, d\textbf{Z} \,d\Gamma \,d\alpha }.      
\end{align}}\vspace{-0.3cm}

\textcolor{black}{To address this challenge, this paper employs Variational Bayes inference (VBI) to derive approximate posterior distributions of hidden variables. Specifically, Variational Bayes Expectation Maximization (VBEM) \cite{ ban2019variational} estimates hidden variables and hyperparameters to identify KCCs with excessive variance in the proposed method. VBEM consists of E-step: Variational Bayesian expectation step to estimate hidden variables $\textbf{X}, \textbf{Z}, \Gamma, \alpha$ by approximating the posterior distribution of hidden variables; and M-step: Variational Bayesian maximization step to update hyperparameters $\text{B}$ by maximizing the expected value of the logarithm of the complete likelihood \cite{han2018bayesian}.}

Let $\boldsymbol{\theta}$ be a vector with all hidden variables in the proposed method (i.e., $\boldsymbol{\theta}=(\textbf{X}, \textbf{Z}, \Gamma, \alpha)$. VBI approximates the posterior distribution of $\boldsymbol{\theta}$, denoted as $q(\boldsymbol{\theta})$, by minimizing Kullback-Leibler (KL) divergence between $q(\boldsymbol{\theta})$ and the true posterior distribution, namely, $p(\boldsymbol{\theta}|\textbf{Y})$ (i.e., $D_{KL} (q(\boldsymbol{\theta})||p(\boldsymbol{\theta}|\textbf{Y})$)) \cite{bishop2006pattern}. $q(\boldsymbol{\theta})$ is factorized as 
\begin{equation*}
    q(\boldsymbol{\theta})=q(\textbf{X})q(\textbf{Z})q(\Gamma)q(\alpha)
\end{equation*}
by the mean-field approximation \cite{cohn2010mean}. \textcolor{black}{The approximate posterior distribution $q(\theta_{i})$, where $\theta_{i}$ is the $i^{th}$ element in the set $\boldsymbol{\theta}$ is derived as follows by minimizing the $D_{KL} (q(\boldsymbol{\theta})||p(\boldsymbol{\theta}|\textbf{Y})$ under the mean-field approximation. 
\begin{equation}\label{eq:12}
\ln{q(\theta_{i})} = \mathbb{E}[\ln{p(\boldsymbol{y},\boldsymbol{\theta}})]_{\boldsymbol{\theta} \setminus \theta_{i}}+const,
\end{equation}
where $\mathbb{E}_{\boldsymbol{\theta} \setminus \theta_{i}}$ denotes the expectation taken with the set $\boldsymbol{\theta}$ without $\theta_{i}$.} \textit{const} can be obtained through normalization. Eq.~\eqref{eq:12} is used in the following E-step of VBEM to approximate the posterior distributions of hidden variables.

\textbf{E-step of VBEM}: The posterior distributions of hidden variables that are related to the KCCs $(\textbf{X})$, the variance of the KCCs $(\Gamma)$, the variable to cluster the KPCs samples $(\textbf{Z})$, and measurement noise from KPCs sample $(\alpha)$ are approximated by Eq.~\eqref{eq:12}, respectively, as follows.
\vspace{-0.0cm}\begin{align}\label{eq:13}
      \ln{q(\textbf{X})} &= \mathbb{E}[ \ln{p(\textbf{Y},\textbf{X}, \textbf{Z}, \Gamma, \alpha)}]_{q(\textbf{Z})q(\Gamma)q(\alpha)} +const  \nonumber \\
      & =\mathbb{E}[ \ln{p(\textbf{Y}|\textbf{X},\alpha})p(\textbf{X}|\textbf{Z},\Gamma;\textbf{B}) ]_{q(\textbf{Z})q(\Gamma)q(\alpha)}+const,
\end{align}
\begin{align}\label{eq:14}
      \ln{q(\Gamma)} &= \mathbb{E}[ \ln{p(\textbf{Y},\textbf{X}, \textbf{Z}, \Gamma, \alpha)} ]_{q(\textbf{X})q(\textbf{Z})q(\alpha)}  +const  \nonumber \\
      & =\mathbb{E}[ \ln{p(\textbf{X}|\textbf{Z},\Gamma;\textbf{B})p(\Gamma \vert a,b)} ]_{q(\textbf{X})q(\textbf{Z})}+const,
\end{align}
\begin{align}\label{eq:15}
 \hspace{-1cm}     \ln{q(\textbf{Z})} &= \mathbb{E}[ \ln{p(\textbf{Y},\textbf{X}, \textbf{Z}, \Gamma, \alpha)} ]_{q(\textbf{X})q(\Gamma)q(\alpha)}  +const  \nonumber \\
      & =\mathbb{E}[ \ln{p(\textbf{X}|\textbf{Z},\Gamma;\textbf{B})} ]_{q(\textbf{X})q(\Gamma)}+const,
\end{align}
\begin{align}\label{eq:16}
   \hspace{-1cm}     \ln{q(\alpha)} &= \mathbb{E}[ \ln{p(\textbf{Y},\textbf{X}, \textbf{Z}, \Gamma, \alpha)} ]_{q(\textbf{X})q(\textbf{Z})q(\Gamma)}  +const  \nonumber \\
      & =\mathbb{E}[ \ln{p(\textbf{Y}|\textbf{X},\alpha)p(\alpha|a,b)} ]_{q(\textbf{X})}+const.
\end{align}
 Based on the statistical inference, the posterior distributions of hidden variables can be derived as
\begin{equation}\label{eq:17}
    q(\textbf{X})=\prod_{k=1}^{K}N(\textbf{x}_{k}|\mu_{k},\Sigma_{k}),
\end{equation}
\begin{equation}\label{eq:18}
    q(\Gamma)=\prod_{g=1}^{G}\prod_{r=1}^{R}\text{Gamma}(\gamma_{g,r}|a_{g,r}, b_{g,r}),
\end{equation}
\begin{equation}\label{eq:19}
    q(\textbf{Z})=\prod_{k=1}^{K}\prod_{g=1}^{G}=\frac{\text{exp}(z_{k,g}\xi_{k,g})}{\sum_{g=1}^{G}\text{exp}(z_{k,g}\xi_{k,g})},
\end{equation}
\begin{equation}\label{eq:20}
    q(\alpha)=\text{Gamma}(\alpha|a_{\alpha}, b_{\alpha}).
\end{equation}
The expectations and moments of distributions in Eqs. \eqref{eq:17},~\eqref{eq:18},~\eqref{eq:19}, and~\eqref{eq:20} are
\vspace{-0.0cm}\begin{equation}\label{eq:21}
    \mu_{k}=\mathbb{E}[\alpha]\Sigma_{k}\Phi^{\top}\textbf{y}_{k},
\end{equation}
\begin{align}\label{eq:22}
\Sigma_{k}&=(\mathbb{E}[\alpha]\Phi^{\top}\Phi+ \nonumber\\
&\text{bdiag}[\sum_{g=1}^{G} \mathbb{E}[ \gamma_{g,1} ] \mathbb{E}[ z_{k,g} ] \text{B}_{1},...,\sum_{g=1}^{G} \mathbb{E}[ \gamma_{g,\text{R}} ] \mathbb{E}[ z_{k,g} ] \text{B}_{\text{R}} ])^{-1},
\end{align}
\begin{align}\label{eq:23}
    \mathbb{E}[ \gamma_{g,r}]=\frac{a_{g,r}}{b_{g,r}} & \nonumber\\
    & \hspace{-1.8cm}= \frac{2a-1 +\sum_{k=1}^{K}\text{ncol}(\text{B}_{r})\mathbb{E}[ z_{k,g} ] }{2b+\sum_{k=1}^{K}\mathbb{E}[ z_{k,g} ](\text{Tr}(\text{B}_{r}(\Sigma_{k,r}+\mu_{k,r}^{\top}\mu_{k,r})))},
\end{align}
\begin{equation}\label{eq:24}
   \hspace{-2.1cm} \mathbb{E}[ z_{k,g} ]=q(z_{k,g}=1)=\frac{\text{exp}(\xi_{k,g})}{\sum_{g=1}^{G}\text{exp}(\xi_{k,g})},
\end{equation}
\begin{equation}\label{eq:25}
    \mathbb{E}[ \alpha ] =  \frac{a_{\alpha}}{b_{\alpha}}=\frac{a+\frac{KM}{2}}{b+\frac{1}{2}\Sigma_{k=1}^{K}(\lVert \textbf{y}_{k}-\Phi \mu_{k}\rVert_{2}^{2}+\text{Tr}(\Phi \Sigma_{k} \Phi^{\top}))},
\end{equation}
where $ \xi_{k,g}=\sum_{r=1}^{R}\textcolor{black}{\text{ncol}}(\text{B}_{r})(\Psi(a_{g,r})-\ln{(b_{g,r})})+\ln(\text{det}(\text{B}_{r}))-\gamma_{g,r}(\mu_{k,r}^{\top}\text{B}_{r}\mu_{k,r}+\textcolor{black}{\text{Tr}}(\text{B}_{r}\Sigma_{k,r}))$. ncol, Tr,  bdiag and $\Psi$ denote the number of columns, trace, block diagonal matrix, and the digamma function \cite{wang2015novel}, respectively. In addition, $\mu_{k,r}$ and $\Sigma_{k,r}$ indicate the posterior mean and the variance of the $r^{th}$ KCCs of $k^{th}$ KPCs sample. Detailed derivations of Eqs.~\eqref{eq:17},~\eqref{eq:18}~\eqref{eq:19}, and~\eqref{eq:20} are provided in the Appendices~\ref{app:app1},~\ref{app:app2}, ~\ref{app:app3}, and~\ref{app:app4}, respectively.
\textbf{M-step of VBEM}: Spatial correlations between KCCs (\textbf{B}) are estimated in this step. Let $\textbf{B}^{OLD}$ as hyperparameter $\textbf{B}$ be updated in the past iteration. Posterior distributions of $\textbf{X}, \textbf{Z}, \Gamma$, and $\boldsymbol{\alpha}$ obtained in Eqs.~\eqref{eq:17},~\eqref{eq:18}~\eqref{eq:19}, and~\eqref{eq:20} are denoted as $q(\textbf{X};\textbf{B}^{OLD}), q(\textbf{Z};\textbf{B}^{OLD}), q(\Gamma;\textbf{B}^{OLD})$, and $q(\alpha;\textbf{B}^{OLD})$ respectively. Then, $\textbf{B}$ can be updated (i.e., $\textbf{B}^{NEW}$) by maximizing the complete likelihood as follows:
\begin{align}
       \textbf{B}^{NEW} &  \nonumber\\
    & \hspace{-1.0cm}  = \argmaxB_{\textbf{B}}  \nonumber \\
    &\hspace{-1.0cm}  \mathbb{E}[ \ln{p(\textbf{Y},\textbf{X},\textbf{Z},\Gamma,\boldsymbol{\alpha};\textbf{B})} ]_{q(\textbf{X};\textbf{B}^{OLD})q(\textbf{Z};\textbf{B}^{OLD})q(\Gamma;\textbf{B}^{OLD})q(\alpha;\textbf{B}^{OLD})}\nonumber\\
       & \hspace{-1.0cm} = \argmaxB_{\textbf{B}}\mathbb{E}[ \ln{p(\textbf{X}|\textbf{Z},\Gamma;\textbf{B})} ]_{q(\textbf{X};\textbf{B}^{OLD})q(\textbf{Z};\textbf{B}^{OLD})q(\Gamma;\textbf{B}^{OLD})}.\label{eq:26}
\end{align}

The correlation matrix of $i^{th}$ correlated list is estimated by maximizing the Eq.~\eqref{eq:26} and taking an inverse  as follows:
\begin{align}\label{eq:27}
    \text{B}_{i}^{-1} 
    =\frac{\sum_{k=1}^{K}\sum_{g=1}^{G}\mathbb{E}[ z_{k,g} ]\mathbb{E}[\gamma_{g,i}](\mu_{k,i}^{\top}\mu_{k,i} +\Sigma_{k,i} )   }{\sum_{k=1}^{K}\sum_{g=1}^{G}\mathbb{E}[ z_{k,g} ] }.
\end{align}
\textcolor{black}{However, as mentioned in Eq.~\eqref{eq:7}, there is a required structure for matrix $\text{B}_{i}^{-1}$ $(i=1,..,r)$ in the multistation assembly system \cite{kong2006mode}, which should be considered in the proposed method. Therefore, $\text{B}_{i}^{-1}$ in Eq.~\eqref{eq:27} can be approximated as}
\begin{equation}\label{eq:28}
  \tilde{\text{B}}_{i}^{-1}=
  \begin{bmatrix}
   1  & \tilde{k}_{i} & \cdots & \tilde{k}_{i} \\
   \tilde{k}_{i}  & 1& \cdots & \vdots \\
   \vdots  & \tilde{k}_{i} & \ddots & \tilde{k}_{i} \\
   \tilde{k}_{i}  & \cdots &\tilde{k}_{i} & 1 
  \end{bmatrix}_{d_{i}\times d_{i}}
\end{equation}
\textcolor{black}{,with $\tilde{k}_{i}=\theta_{1}^{i}/\theta_{0}^{i}$, where $\theta_{0}^{i}$ and $\theta_{1}^{i}$ are the average on the diagonal and main sub-diagonal elements of matrix $\text{B}_{i}^{-1}$ in Eq.~\eqref{eq:27}, respectively \cite{bastani2018fault,zhang2013extension}. The detailed derivation of Eq.~\eqref{eq:27} is described in Appendix~\ref{app:app5}.}

Algorithm \ref{alg:alg1} shows the procedure of the proposed CSSBL method. Given the measurement samples of KPCs ($\textbf{Y}$) and fault pattern matrix ($\Phi$), the proposed method estimates the following variables and parameters in E and M steps, respectively.
\vspace{-0.0cm}\begin{itemize}
    \item E-step: KCCs $(\textbf{X})$, the variance of the KCCs $(\Gamma)$, the variable to cluster the KPCs samples $(\textbf{Z})$, and measurement noise from KPCs $(\alpha)$.\vspace{-0.0cm}
    \item M-step: Spatial correlations of KCCs ($\textbf{B}$).
\end{itemize}\vspace{-0.0cm}
\textcolor{black}{These steps iterate until the estimator of KCCs ($\mu$) \textcolor{black}{converges}, namely, $\lVert \mu^{t-1}-\mu^{t} \rVert_{\infty}<s$, where $\lVert \cdot
 \rVert_{\infty}$ indicates infinity norm and $s$  is a user-defined threshold (e.g., $s=10^{-6}$). The method also terminates if it reaches the maximum number of iterations ($\mathbf{T}$).} Finally, $\Sigma_{k}$ in Eq.~\eqref{eq:22}  provides the variance of KCCs in the $k^{th}$ KPCs sample.  

\begin{algorithm}[!htb]\label{alg:alg1}
    \textbf{Input}: Measurement Samples of KPCs ($\textbf{Y}$), Fault pattern matrix ($\Phi$), Threshold ($\gamma$).\\
    \textbf{Set}\hspace{0.25em}$a=b=10^{-4}$.\\
    \textbf{Initialize}\hspace{0.25em}$\text{B}=\text{bdiag}[\textbf{I}_{1},...,\textbf{I}_{\text{R}}], \alpha=1,  t=1.$\\
    \textbf{While}  $\lVert \mu^{t-1}-\mu^{t} \rVert_{\infty}\geq \gamma $ or $t\leq \mathbf{T}$ do\\
    \textbf{E-step of VBEM}:\\
 \indent   \hspace{5.00em} Update $ \mu$ using Eq.~\eqref{eq:21}\\
 \indent   \hspace{5.00em} Update $ \Sigma$ using Eq.~\eqref{eq:22}\\ 
  \indent  \hspace{5.00em}  Update $ \Gamma$ using Eq.~\eqref{eq:23}\\
   \indent   \hspace{5.00em} Update $\textbf{Z}$ using Eq.~\eqref{eq:24}\\
   \indent   \hspace{5.00em} Update $\alpha$ using Eq.~\eqref{eq:25}\\   
    \hspace{2.50em} \textbf{M-step of VBEM}:\\
 \indent      \hspace{5.00em} Update $\textbf{B}$ using Eq.~\eqref{eq:28}\\
 \indent    \hspace{2.50em} $t=t+1$\\
    \textbf{End}\\
    \textbf{Output}:  Variance of KCCs (Eq.~\eqref{eq:22}).
    \caption{Proposed CSSBL method}
\end{algorithm}\vspace{-0.35cm}

\vspace{-0.3cm}
\section{Numerical Case Studies} \label{s:sec4}
This section provides numerical studies to compare the performance between the proposed method and benchmark methods. \textcolor{black}{The studies evaluate the performance by varying the correlations ($k_{i}$ in Eq.~\eqref{eq:7}) between the KCCs in nonstationary process faults.}
 All the numerical case studies consist of 20 independent trials. The code of the proposed algorithm is implemented in Matlab 2021. The CPU used in case studies is an Intel\textsuperscript{\textregistered} Core\textsuperscript{\texttrademark} Processor i7-8750H.


This section compares the proposed method with the following benchmark methods.
\vspace{-0.0cm}\begin{itemize}
\item User Grouping Sparse Bayesian Learning (i.e., UGSBL) proposed in \cite{dai2018joint} is an SBL method that clusters the KPCs samples into groups sharing the same process faults in sparse estimation.
\item \textcolor{black}{Spatially Correlated Bayesian Learning (i.e., SCBL) proposed in \cite{bastani2018fault} is an SBL method to diagnose process faults by exploiting the spatial correlation of KCCs. }  
\item \cite{lee2020variation} proposed the SBL method that considers prior knowledge of the process faults in sparse estimation. 
\item \textcolor{black}{MSBL proposed in \cite{wipf2007empirical} is a basic SBL method that assumes the independence of KCCs and stationarity of process faults.}
\end{itemize}


\noindent\textbf{Data Generations}: The KCCs from $k^{th}$ KPCs sample ($ \textbf{x}_{k}$) are generated based on the following procedure. $ \textbf{x}_{k}$ consists of the correlated KCCs list and independent KCCs.
Given the $\text{B}_{i}^{-1}$ from Eq.~\eqref{eq:7}, the $i^{th}$ correlated KCCs list from $k^{th}$ sample ($ \textbf{x}_{k,i}$) is generated by    
\begin{equation}\label{eq:29}
    \textbf{x}_{k,i}=(\textbf{u}^{\top}\text{chol}((\gamma_{i}\text{B}_{i})^{-1}))^{\top},
\end{equation}
where $\textbf{u}\in \mathbb{R}^{3 \times 1}$ is generated from $i.i.d.$ standard Gaussian distribution. chol($\cdot$) is defined as the Cholesky decomposition operator \cite{trefethen1997numerical}. \textcolor{black}{The independent KCCs are also generated from Eq.~\eqref{eq:29} with $\text{B}_{i}=\textbf{I}_ {1\times1}$.} To provide the process faults whose KCCs have excessive variance, $\gamma_{i}^{-1}$ in Eq.~\eqref{eq:29} is provided as one. In contrast, $\gamma_{i}^{-1}$ equals 0.01 for KCCs with non-process fault since the variation of KCCs always exists \cite{bastani2016compressive}. 
To generate the corresponding $k^{th}$ KPCs sample ($\textbf{y}_{k}$), a dictionary matrix $\Phi \in \mathbb{R}^{M \times N}$ is constructed with columns drawn from the surface of a unit hyper-sphere uniformly \cite{donoho2006most}. The measurement errors ($\textbf{v}_{k}$) are generated from the Gaussian distribution with a variance of $10^{-6}$.
Finally, the measurements of $k^{th}$ KPCs sample is built by $ \textbf{y}_{k} =\Phi \textbf{x}_{k}+ \textbf{v}_{k}$. 

\begin{figure}[h]
    \centering
    \includegraphics[width=0.5\textwidth]{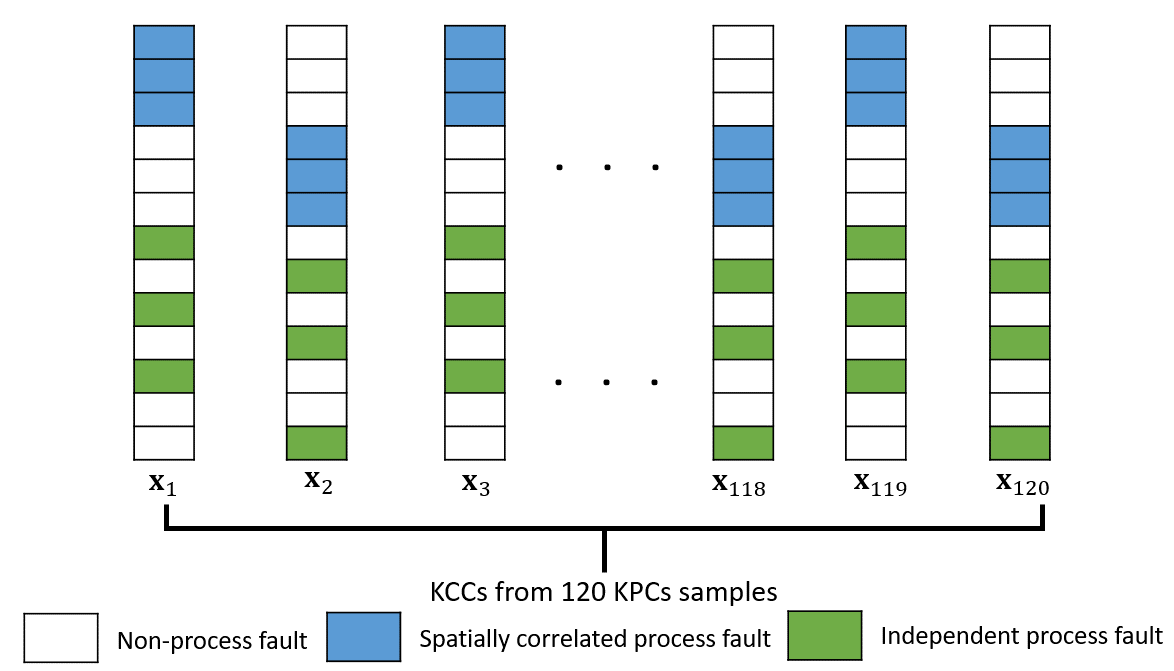}\vspace{-0.0cm}
    \caption{\textcolor{black}{Data generation of KCCs ($\textbf{X}=[\textbf{x}_{1},...,\textbf{x}_{120}]$ in Eq.~\eqref{eq:3}) for the 120 KPCs samples  with spatially correlated KCCs and nonstationary process faults.}}
    \label{fig:fig6}\vspace{-0.7cm}
\end{figure}

To provide the underdetermined system in the numerical studies, $M$ and $N$ in Eq.~\eqref{eq:3} are set as 8 and 40, respectively. In addition,  two lists of correlated KCCs are generated with the size of three, respectively (i.e., $r=2$ and $d_{1}=d_{2}=3$ in Eqs.~\eqref{eq:6} and~\eqref{eq:7}). The correlation matrix $\text{B}_{i}^{-1}\in \mathbb{R}^{3\times 3}, (i=1, 2)$ are defined based on Eq.~\eqref{eq:7}. Assume coefficients $k_{1}$ and $k_{2}$ equal  for convenience. The coefficients are varied in the case studies.  For example, the correlation coefficient 0.1 represents a case with a low correlation between the correlated KCCs, while 0.9 illustrates a case with a high correlation. 
\textcolor{black}{To generate nonstationary process faults along the KPCs samples, two groups ($G=2$ in Eq.~\eqref{equation:7}) of process faults are generated as shown in Fig.~\ref{fig:fig6}. \textcolor{black}{ Both Groups 1 and 2 have  one list of correlated and three independent process faults, respectively. However, both groups do not have any common process faults.}  As shown in Fig.~\ref{fig:fig6}, 60 KCCs are generated from each group and randomly ordered. Based on these KCCs, 120 KPCs samples are generated and provided for analysis.}


\vspace{-0.0cm}\begin{table*}[t]
\centering
\caption{Performance of AUC in various correlations between correlated KCCs ($k_{1}, k_{2}$) with nonstationary process faults (G=2) in the numerical studies.}
\label{tab: table1}
\begin{adjustbox}{width=0.7\textwidth}\begin{tabular}{ccccccccccc}
\hline\hline
 $k_{1}=k_{2}$   & 0.1  & 0.2  & 0.3  & 0.4  & 0.5 & 0.6  & 0.7  & 0.8  & 0.9  & 0.95  \\
\hline MSBL   & 0.91 & 0.91 & 0.90 &  0.90 & 0.89  & 0.88  & 0.87 &0.82 &0.79 &0.79  \\
\hline SSBL     & 0.91  &0.91  & 0.91 &0.91 & 0.91 & 0.92 &0.92  &0.89  &0.90  &0.90   \\
\hline UGSBL   & 0.95 & 0.92 & 0.94 &0.94 &0.91 & 0.90 & 0.90 & 0.87 & 0.84 & 0.84  \\ 
\hline \cite{lee2020variation}   & 0.89 & 0.87  & 0.87  & 0.89  &  0.84  & 0.85 &  0.82 & 0.79 & 0.76 & 0.76  \\
\hline
\begin{tabular}[c]{@{}c@{}}\textbf{CSSBL}\\ \vspace{-0.0cm}\textbf{(Proposed)}\end{tabular} & \textbf{0.95} & \textbf{0.93} & \textbf{0.95} & \textbf{0.95} & \textbf{0.94} & \textbf{0.94} & \textbf{0.95} & \textbf{0.95} & \textbf{0.95} & \textbf{0.97}  \\ \hline \hline
\end{tabular}
\end{adjustbox}\vspace{-0.4cm}
\end{table*}

\vspace{-0.0cm}\vspace{-0.0cm}
\noindent \textbf{Performance Evaluation}: To evaluate the effectiveness of the proposed method, two performance measures are used in this paper. The first measure evaluates the process fault detection capability, and the second measure is related to estimation accuracy. Each measure is evaluated by varying the correlation coefficients of spatially correlated KCCs.
\vspace{-0.0cm}\begin{table*}[t]
\centering
\caption{Performance of NMSE in various correlations between correlated KCCs ($k_{1}, k_{2}$) with nonstationary process faults (G=2) in the numerical studies.}
\label{tab: table2}
\begin{adjustbox}{width=0.7\textwidth}\begin{tabular}{ccccccccccc}
\hline\hline
 $k_{1}=k_{2}$   & 0.1  & 0.2  & 0.3  & 0.4  & 0.5 & 0.6  & 0.7  & 0.8  & 0.9  & 0.95  \\
\hline MSBL   & 0.54 & 0.56 & 0.59& 0.57  & 0.64   & 0.67   & 0.68  & 0.82 & 0.88 & 0.88  \\
\hline SSBL     & 0.62  &0.68  & 0.66 &0.63 & 0.66 & 0.63 &0.66  &0.76  &0.73  &0.78   \\
\hline UGSBL   & \textbf{0.25} & 0.32 & 0.31 & 0.28 &0.45 & 0.49 & 0.48 & 0.78 & 0.97 & 0.93  \\ 
\hline \cite{lee2020variation}   & 0.61 & 0.65 & 0.67  & 0.61  & 0.74 & 0.75  & 0.74  & 0.87  & 0.95  & 0.91 \\
\hline
\begin{tabular}[c]{@{}c@{}}\textbf{CSSBL}\\ \vspace{-0.0cm}\textbf{(Proposed)}\end{tabular} & 0.26 & \textbf{0.28} & \textbf{0.27} & \textbf{0.23} & \textbf{0.28} & \textbf{0.27} & \textbf{0.23} & \textbf{0.22} & \textbf{0.21} & \textbf{0.13}  \\ \hline \hline
\end{tabular}
\end{adjustbox}\vspace{-0.5cm}
\end{table*}

\noindent \textit{1) Fault Detection Capability:} 
First, the proposed method requires correctly identifying process faults whose KCCs have excessive variance. In order to assess the capability of detecting process faults under sparse conditions, the area under the receiver operating characteristics (ROCs) curve (AUC) is utilized  \cite{hanley1982meaning}. AUC  is widely used for evaluating the accuracy of binary classification. Specifically, AUC measures the quality of the classification accuracy between process faults and non-process faults irrespective of the threshold of variance. In other words, AUC accounts for the trade-off between the type 1 and type 2 errors. 
When the method is capable of achieving perfect classification results, the resulting AUC value will be 1. However, the AUC with a value close to 0.5 denotes that the estimated variances of KCCs from the method are similar between the process faults and non-process faults.

Table~\ref{tab: table1} represents that the proposed method always achieves the best performance of AUC in the various spatial correlation between KCCs. Specifically, the AUC of the proposed method is close to 1.0 at every correlations levels, representing the perfect classification between the process faults and non-process faults in terms of estimated variance. Since UGSBL clusters the KPCs samples that share the same process faults, the method shows a comparable result to the proposed method when the spatial correlation of KCCs is relatively low and moderate. However, the performance degrades in the high spatial correlation since the method assumes independence between all KCCs. The performance of SSBL is robust to correlations. This is because SSBL utilizes the correlation of correlated KCCs in its estimation. However,  SSBL performs worse than the proposed method since this method assumes the same process faults between all KPCs samples, which are unsuitable for dealing with nonstationary process faults. MSBL and  \cite{lee2020variation} show poor performances since these methods assume the independence of KCCs and stationarity of process faults.



\noindent \textit{2) Estimation Accuracy:} In addition to correctly identifying the process fault, the method needs to accurately estimate the variance of process faults. After the process faults are identified, the practitioners are interested in how large the variances are to make decisions for the maintenance of the process. To achieve this objective, this paper provides the normalized mean squared error (NMSE) between true and estimated variance.

Table~\ref{tab: table2} represents that the proposed method  achieves the best performance in the estimation accuracy of the variance in most of the correlations. 
It implies that the proposed method is useful not only for the process faults identification but also for measuring how severe the process faults are. Compared to benchmark methods, the effectiveness of the proposed method in the NMSE results is noticeable in the presence of high spatial correlations between KCCs.
Unlike AUC performance, the NMSE of SSBL increases as the correlation increases. It implies that SSBL successfully identifies the process faults utilizing the correlation information but cannot accurately estimate the variance since SSBL assumes the stationary process faults over KPC samples. The performance of UGSBL, MSBL, and \cite{lee2020variation} represent similar trends to the performance of AUC.

\vspace{-0.3cm}
\section{Real-World Simulation Case Studies}\label{s:sec5}

\begin{figure}[!htbp]
    \centering
        \includegraphics[width=0.5\textwidth]{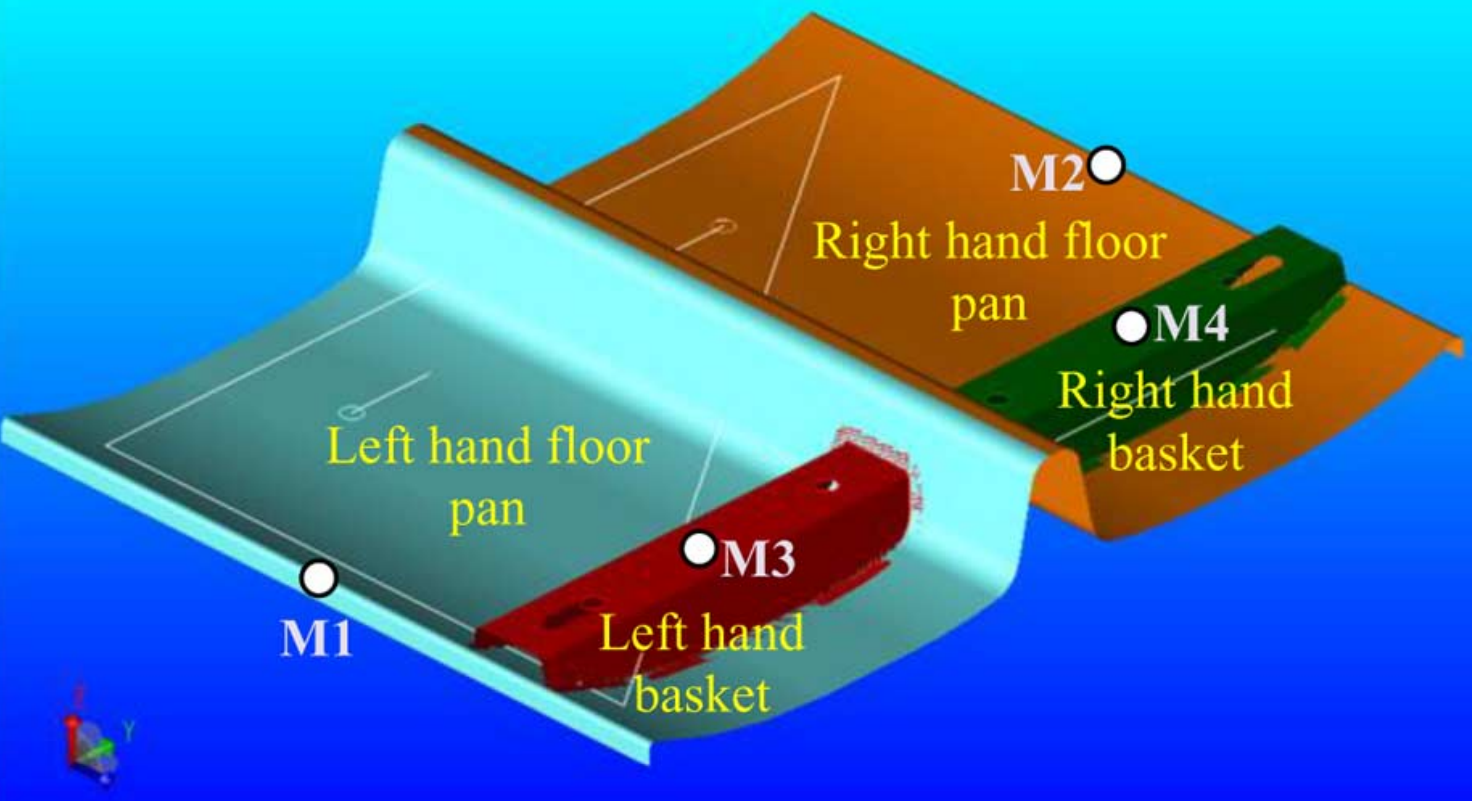}\vspace{-0.0cm}
    \caption{Floor-pan assembly model \cite{bastani2012fault}.}
    \label{fig:fig8}\vspace{-0.6cm}
\end{figure}

\begin{figure}[!htbp]
    \centering
    \includegraphics[width=0.5\textwidth]{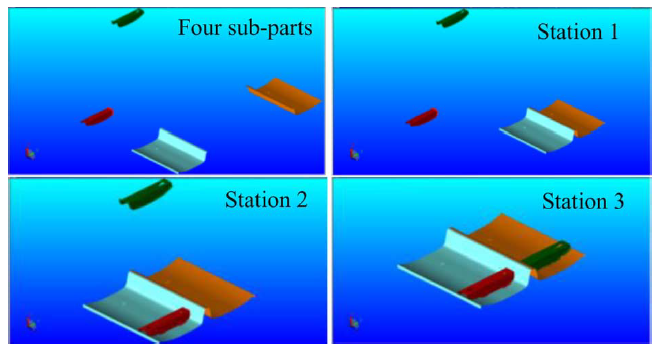}\vspace{-0.0cm}
    \caption{Floor-pan assembly model from three assembly stations. \cite{bastani2012fault}.}
    \label{fig:fig9}\vspace{-0.1cm}
\end{figure}

An assembly operation from an actual auto-body assembly process is used as a real-world case study. Fig.~\ref{fig:fig8} describes a car's floor pan, which is the assembled product from this process. The assembled product consists of four parts, including the right bracket, left bracket, right floor pan, and left floor pan.
Fig.~\ref{fig:fig9} shows the process assembly procedure consisting of three stations. During the assembly process, the parts are held by fixtures, which are the KCCs in this process \cite{bastani2016compressive}. KPCs are measured from four points, namely, M1, M2, M3, and M4, respectively, as shown in Fig.~\ref{fig:fig8}. Every part has a designated location for measuring the KPCs. For instance, part 1 has M1, part 2 has M2, part 3 has M3, and part 4 has M4. These measurements can be taken at each station once the relevant part has been assembled in the preceding stations. For example, M3 on part 3 cannot be measured in station 1 because part 3 has not yet been assembled at station 1 \cite{bastani2016compressive}. 
In the designated location, KPCs are measured in three directions (X, Y, and Z). In this assembly process, there exists a total of 33 KCCs, which are dimensional errors of fixture locators \cite{bastani2018fault}. The fault pattern matrix $\Phi$ is established based on the literature \cite{bastani2016compressive,huang2007stream,kong2009variation} and provided in Appendix \ref{app:app6}. Since the number of measurements (12) is less than that of KCCs (33), it causes an underdetermined system in the fault quality linear model. Therefore, sparse estimation is required to identify process faults.
\vspace{-0.3cm}

\vspace{-0.3cm}
\textcolor{black}{In this assembly process, two lists of correlated KCCs exist; the rest are assumed to be independent \cite{bastani2018fault}. The first list comprises six KCCs, including KCC8, KCC9, KCC10, KCC11, KCC12, and KCC13. The second list is composed of KCC31, KCC32, and KCC33. The correlation structures of the two groups are described (for simplification, the lower triangular terms are removed) as follows \cite{bastani2018fault}: }

\vspace{-0.0cm}\noindent\begin{tabularx}{0.5\textwidth}{@{}XX@{}}
  \begin{equation*}
    {\text{B}}_{1}^{-1}=
  \begin{bmatrix}
   1  & {k}_{1} & \cdots & {k}_{1} \\
     & 1& \cdots & \vdots \\
     &  & \ddots & {k}_{1} \\
    &  & & 1 
  \end{bmatrix}_{6 \times 6},
    \label{equation:10}
  \end{equation*} &
  \begin{equation*}
  {\text{B}}_{2}^{-1}=
  \begin{bmatrix}
   1  & {k}_{2} & {k}_{2} \\
     & 1&  {k}_{2} \\
   &   & 1 \\
  \end{bmatrix}_{3 \times 3}.
    \label{equation:11}
  \end{equation*} 
\end{tabularx}
\textcolor{black}{To provide nonstationary process faults along the KPCs samples collected over time, two groups are provided. \textcolor{black}{In the first group, the first correlated list is provided as process faults. For the process faults of the second group, all the correlated lists and one independent KCCs are included.} Therefore, the case study represents the situation when the process faults evolve over time since the variation of certain KCCs will be propagated to other KCCs if the process faults are not mitigated immediately during the process. 50 KPCs samples are generated from each group. The variances of process faults, non-process faults, and noise of KPCs are determined to be the same value as in Section~\ref{s:sec4}. In addition, performance evaluation measures and benchmark methods used in Section~\ref{s:sec4} are still utilized in Section~\ref{s:sec5}.} Table~\ref{tab: table3} shows the performance comparison in AUC and NMSE by varying the correlation coefficients. 
For convenience, the correlation coefficients $k_{1}$ and $k_{2}$ are equal and are selected from $\{0.1, 0.3, 0.6, 0.9, 0.95 \}$ in the case studies.
Specifically, the AUC of the proposed method is close to 1.0 at every correlations levels, representing the perfect classification between the process faults and non-process faults in terms of estimated variance. The effectiveness of the proposed method is significant in NMSE when a high correlation exists. Specifically, the NMSE of the proposed method is less than 10\% of that of all the benchmark methods when the correlation coefficient is 0.9 and 0.95.

\vspace{-0.0cm}\begin{table*}[t]
\centering
\caption{Performances in various correlations between KCCs ($k_{1}, k_{2}$) in the actual correlated locators' list with nonstationary process faults (G=2).}
\label{tab: table3}
 \begin{adjustbox}{width=0.7\textwidth}\begin{tabular}{cccccc|ccccc}
\hline\hline
       & \multicolumn{5}{c}{AUC} & \multicolumn{5}{c}{NMSE}    \\ 
\cline{2-11}          
 $k_{1}=k_{2}$   & 0.1  & 0.3  & 0.6  & 0.9  & 0.95 & 0.1  & 0.3  & 0.6  & 0.9  & 0.95 \\
\hline MSBL   & 0.96 & 0.92 & 0.89 &  0.72 &  0.67 &0.46  &0.84  & 1.91 & 3.05 & 3.32 \\
\hline SSBL     & 0.95 & 0.91 & 0.94 & 0.94 & 0.99 & 4.26 & 12.84 & 11.97 & 30.32 & 1.60\\
\hline UGSBL   & 0.98 & 0.97 & 0.95 & 0.90 & 0.89  & \textbf{0.29} & 0.54 & 1.68 & 3.09 & 3.30  \\ 
\hline \cite{lee2020variation}   &0.94 & 0.92 & 0.89 & 0.86 & 0.87  & 0.47 & 0.59 & 0.80 & 1.05 & 1.10  \\
\hline
\begin{tabular}[c]{@{}c@{}}\textbf{CSSBL}\\ \vspace{-0.0cm}\textbf{(Proposed)}\end{tabular} & \textbf{0.99} & \textbf{0.99} & \textbf{1.00} & \textbf{1.00} & \textbf{1.00} & 0.39 & \textbf{0.27} & \textbf{0.17} & \textbf{0.10} & \textbf{0.09}  \\ \hline \hline
\end{tabular}
\end{adjustbox}
\end{table*}


\vspace{-0.0cm}
\begin{figure*}[t]\vspace{-0.4cm}
   \hspace{1.5cm}  {\includegraphics[width=0.8\textwidth]{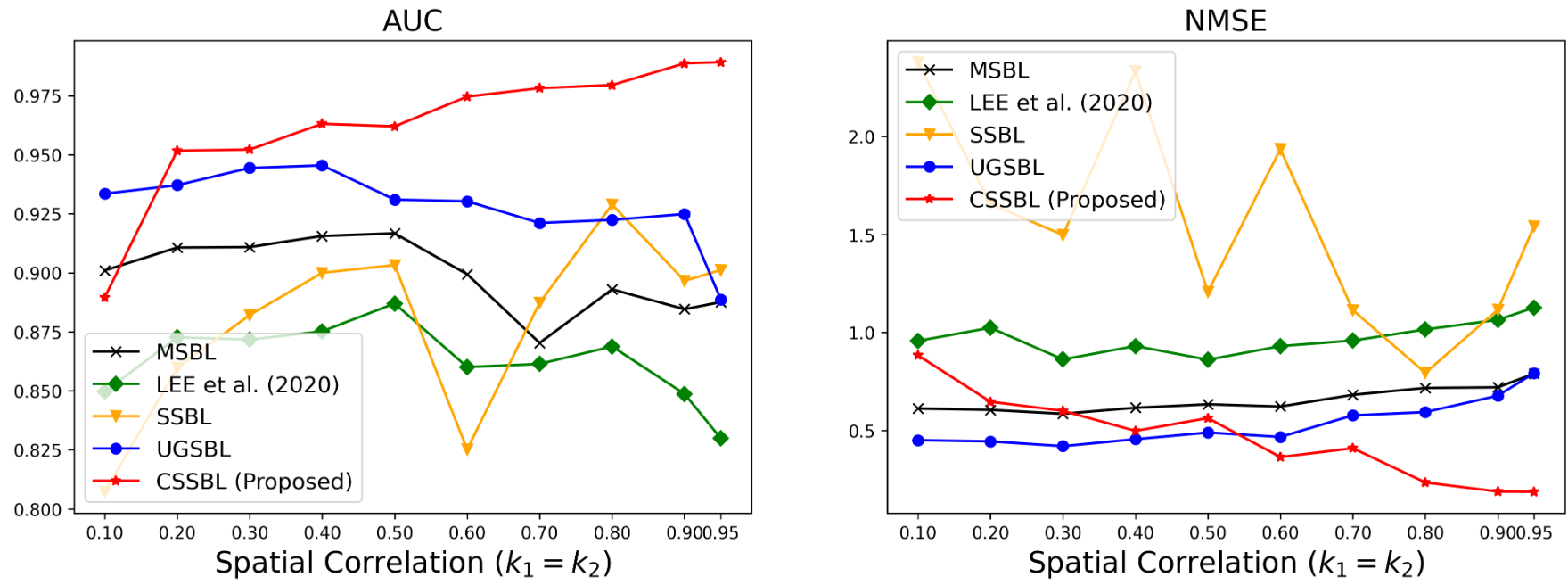}}\vspace{-0.0cm}
    \caption{Performances in various correlations between KCCs ($k_{1}, k_{2}$) in the randomly selected correlated locators' list with nonstationary process faults (G=2).}
    \label{fig:fig10}\vspace{-0.5cm}
\end{figure*}\vspace{-0cm}



To demonstrate the generalizability of the proposed method, the correlated KCCs are randomly selected in further case studies. Assuming there exist two correlated lists with the size of three, respectively. Therefore,  six KCCs are randomly selected from 33 KCCs for two correlated lists in each trial. 
\textcolor{black}{As in the previous study, one correlated list is provided as process faults in the first group, while two correlated lists and two independent KCCs are used as process faults in the remaining second group.} 
Fig.~\ref{fig:fig10} shows the performances of all the methods in AUC and NMSE results in ten different spatial correlations (i.e., $k_{1}=k_{2}$). From Fig.~\ref{fig:fig10} (a), it is prominent that the proposed method successfully differentiates between process faults and non-process faults. Specifically, the AUC of the proposed method is nearly 1.0, while those of most benchmark methods are below 0.9 in the highly correlated case studies. In addition, Fig.~\ref{fig:fig10} (b) represents that the proposed method  provides valuable information to the practitioners to determine when to stop the operations to maintain the process when a high correlation exists between KCCs. All the benchmark methods illustrate similar trends compared to Section~\ref{s:sec4} since the methods cannot consider both the spatial correlation of KCCs and the nonstationary process faults. 
\vspace{-0.3cm}
\section{Conclusions}\label{s:sec6}
This paper proposes a novel sparse hierarchical Bayesian method, CSSBL, to effectively identify the sparse process faults in multistation assembly systems. The method identifies process faults by considering the spatial correlation of KCCs and nonstationary process faults among the multiple KPCs. Since posterior distributions of KCCs in the proposed method are computationally intractable, this paper derives approximate posterior distributions of KCCs via Variational Bayes inference. The proposed method's effectiveness is validated by numerical cases and real-world simulation application using an actual auto-body assembly system. 
\textcolor{black}{In this work, the temporal correlations of the KCCs are not considered. However, a temporal correlation exists among the KCCs because of the degradation of wear of production tooling over time \cite{shi2022process} or the machine-tool thermal distortion \cite{abellan2012state}. Therefore, extending the proposed CSSBL to consider the temporal correlation of KCCs makes the method more realistic in the actual multistation assembly system. Since the proposed method is not designed for specific processes, it can be effectively applied to other domains, including healthcare and communication systems, for their process monitoring.}

\appendices
\section{\emph{Inference for Eq.~\eqref{eq:17} }}\label{app:app1}
Based on Eq.~\eqref{eq:13},
$\ln{q(\textbf{X})} = \mathbb{E}[ \ln{p(\textbf{Y},\textbf{X}, \textbf{Z}, \Gamma, \alpha)} ]_{q(\textbf{Z})q(\Gamma)q(\alpha)}  +const$. Therefore,
\begin{align}
    \ln{q(\textbf{X})} &\propto \mathbb{E}[ \ln{p(\textbf{Y}|\textbf{X},\alpha})p(\textbf{X}|\textbf{Z},\Gamma;\textbf{B}) ]_{q(\textbf{Z})q(\Gamma)q(\alpha)} \nonumber\\
    & \propto \mathbb{E}[ \ln{p(\textbf{Y}|\textbf{X},\alpha)}]_{q(\alpha)} + \mathbb{E}[ \ln{p(\textbf{X}|\textbf{Z},\Gamma;\textbf{B})} ]_{q(\textbf{Z})q(\Gamma)} \nonumber \\
    & \propto \sum_{k=1}^{K} (\mathbb{E}[ \ln{p(\textbf{y}_{k}|\textbf{x}_{k},\alpha)} ]_{q(\boldsymbol{\alpha})}+ \mathbb{E}[ \ln{p(\textbf{x}_{k}|\textbf{Z},\Gamma;\textbf{B})} ]_{q(\textbf{Z})q(\Gamma)}).\label{eq:30}
\end{align} 
Each term in Eq.~\eqref{eq:30} is proportional to as follows:
\begin{align*}
       & \mathbb{E}[ \ln{p(\textbf{y}_{k}|\textbf{x}_{k},\alpha)} ]_{q(\alpha)}+ \mathbb{E}[ \ln{p(\textbf{x}_{k}|\textbf{Z},\Gamma;\textbf{B})} ]_{q(\textbf{Z})q(\Gamma)}\\
       & \propto -\frac{\mathbb{E}[\alpha]}{2} \lVert \textbf{y}_{k}-
       \Phi \textbf{x}_{k} \rVert_{2}^{2}\\
       &-\frac{1}{2}\textbf{x}_{k}^{\top}\text{bdiag}[\sum_{g=1}^{G} \mathbb{E}[ \gamma_{g,1} ] \mathbb{E}[ z_{k,g} ] \text{B}_{1},...,
       \sum_{g=1}^{G} \mathbb{E}[ \gamma_{g,\text{R}} ] \mathbb{E}[ z_{k,g} ] \text{B}_{\text{R}}]\textbf{x}_{k}.
\end{align*}
Therefore, $q(\textbf{x}_{k})$ follows the Gaussian distribution as follows:
\begin{equation*}
    q(\textbf{x}_{k})=N(\textbf{x}_{k}|\mu_{k},\Sigma_{k}),
\end{equation*}
where $\mu_{k}=\mathbb{E}[\alpha]\Sigma_{k}\Phi^{\top}\textbf{y}_{k}$, and \textcolor{black}{$\Sigma_{k}=(\mathbb{E}[\alpha]\Phi^{\top}\Phi+\text{bdiag}[\sum_{g=1}^{G} \mathbb{E}[ \gamma_{g,1} ] \mathbb{E}[ z_{k,g} ] \text{B}_{1},...,\sum_{g=1}^{G} \mathbb{E}[ \gamma_{g,\text{R}} ] \mathbb{E}[ z_{k,g} ] \text{B}_{\text{R}} ])^{-1}$.}

\section{\emph{Inference for Eq.~\eqref{eq:18} }}\label{app:app2}
Based on Eq.~\eqref{eq:14}, $\ln{q(\Gamma)}=\mathbb{E}[ \ln{p(\textbf{Y},\textbf{X}, \textbf{Z}, \Gamma,\alpha}) ]_{q(\textbf{X})q(\textbf{Z})q(\alpha)}+const.$ Therefore,
\begin{align*}
    \ln{q(\Gamma)} & \propto \mathbb{E}[ \ln{p(\textbf{X}|\textbf{Z},\Gamma;\textbf{B})p(\Gamma|a,b) ]}_{q(\textbf{X})q(\textbf{Z})}\\
    &\propto  \mathbb{E}[ \ln{p(\textbf{X}|\textbf{Z},\Gamma;\textbf{B})]}_{q(\textbf{X})q(\textbf{Z})} +\ln{p(\Gamma|a,b)}\\
    & \propto \sum_{g=1}^{G}\sum_{r=1}^{\text{R}}((2a-2)\ln{\gamma_{g,r}-2b\gamma_{g,r}})\\
    &-\sum_{g=1}^{G}\sum_{r=1}^{\text{R}}\sum_{k=1}^{K}\mathbb{E}[z_{k,g}]\gamma_{g,r}\mathbb{E}[ \textbf{x}_{k,r}^{\top}\text{B}_{r}\textbf{x}_{k,r} ] \\
   & +\text{ncol}(\text{B}_{r})\mathbb{E}[z_{k,g}]\ln{\gamma_{g,r}}\\
      \begin{split}
   & \propto \sum_{g=1}^{G}\sum_{r=1}^{\text{R}}(
(2a-1+\sum_{k=1}^{K}\text{ncol}(\text{B}_{r})\mathbb{E}[z_{k,g}]-1)\ln{\gamma_{g,r}} \\ & -(2b+\sum_{k=1}^{K}\mathbb{E}[z_{k,g}]\mathbb{E}[ \textbf{x}_{k,r}^{\top}\text{B}_{r}\textbf{x}_{k,r} ]) {\gamma_{g,r}}) 
    \end{split}\\ 
     \begin{split}
   & \propto \sum_{g=1}^{G}\sum_{r=1}^{\text{R}}(
(2a-1+\sum_{k=1}^{K}\text{ncol}(\text{B}_{r})\mathbb{E}[z_{k,g}]-1)\ln{\gamma_{g,r}} \\ & -(2b+\sum_{k=1}^{K}\mathbb{E}[z_{k,g}](\text{Tr}(\text{B}_{r}\Sigma_{k,r})
  +\mu_{k,r}^{\top}\text{B}_{r}\mu_{k,r})) {\gamma_{g,r}}). 
    \end{split}
\end{align*}
Therefore, $q(\gamma_{g,r})$, follows Gamma distribution as follows:
\begin{equation*}
    q(\gamma_{g,r})\sim \Gamma(\gamma_{g,r}|a_{\gamma_{g,r}},b_{\gamma_{g,r}}),
\end{equation*}
where $a_{\gamma_{g,r}} = 2a-1+\sum_{k=1}^{K}\text{ncol}(\text{B}_{r})\mathbb{E}[z_{k,g}]$ and $b_{\gamma_{g,r}}=2b+\sum_{k=1}^{K}\mathbb{E}[z_{k,g}](\text{Tr}(\text{B}_{r}(\Sigma_{k,r}+\mu_{k,r}^{\top}\mu_{k,r})))$.

\section{\emph{Inference for Eq.~\eqref{eq:19} }}\label{app:app3}
Based on Eq.~\eqref{eq:15}, $\ln{q(\textbf{Z})}=\mathbb{E}[ \ln{p(\textbf{Y},\textbf{X}, \textbf{Z}, \Gamma,\alpha}) ]_{q(\textbf{X})q(\Gamma)q(\alpha)}+const.$ Therefore,
\begin{align*}
    \ln{q(\textbf{Z})} &\propto \mathbb{E}[ \ln{p(\textbf{X}|\textbf{Z},\Gamma; \textbf{B})} ]_{q(\textbf{X})q(\Gamma)} \nonumber\\
    \begin{split}
     & \propto \sum_{k=1}^{K}\sum_{g=1}^{G}z_{k,g}\mathbb{E}[[\ln(\text{det}(\gamma_{g,1}\text{B}_{1})^{\frac{1}{2}}\cdots\text{det}(\gamma_{g,\text{R}}\text{B}_{\text{R}})^{\frac{1}{2}} \\
    & \text{exp}(-\frac{1}{2}\textbf{x}_{k}^{\top} \text{bdiag}[\gamma_{g,1}\text{B}_{1},...,\gamma_{g,\text{R}}\text{B}_{\text{R}}]\textbf{x}_{k})]]_{q(\textbf{X})q(\Gamma)}
    \end{split}     \\
    \begin{split}
     & \propto \sum_{k=1}^{K}\sum_{g=1}^{G}z_{k,g}\mathbb{E}[[\ln(\text{det}(\gamma_{g,1}\text{B}_{1}))\cdots\ln(\text{det}(\gamma_{g,\text{R}}\text{B}_{\text{R}})) \\
     & -(\textbf{x}_{k}^{\top} \text{bdiag}[\gamma_{g,1}\text{B}_{1},...,\gamma_{g,\text{R}}\text{B}_{\text{R}}]\textbf{x}_{k})]]_{q(\textbf{X})q(\Gamma)}  
     \end{split} \\
     \begin{split}
     & \propto \sum_{k=1}^{K}\sum_{g=1}^{G}z_{k,g}\mathbb{E}[\text{ncol}(\text{B}_{1})\ln(\gamma_{g,1})+\ln(\text{det}(\text{B}_{1}))+\cdots \\
     &+\text{ncol}(\text{B}_{\text{R}})\ln(\gamma_{g,\text{R}})+\ln(\text{det}(\text{B}_{\text{R}})) \\
     &-(\textbf{x}_{k}^{\top} \text{bdiag}[\gamma_{g,1}\text{B}_{1},...,\gamma_{g,\text{R}}\text{B}_{\text{R}}]\textbf{x}_{k})]]_{q(\textbf{X})q(\Gamma)}  
     \end{split}    \\
     \begin{split}
     & \propto \sum_{k=1}^{K}\sum_{g=1}^{G}z_{k,g}\sum_{r=1}^{\text{R}}\text{ncol}(\text{B}_{r})(\mathbb{E}[{\ln{\gamma_{g,r}}}])+\ln(\text{det}(\text{B}_{r}))\\
     &-\mathbb{E}[\gamma_{g,r}](\mu_{k,r}^{\top}\text{B}_{r}\mu_{k,r}+\text{Tr}(\text{B}_{r}\Sigma_{k,r})).
     \end{split}        
\end{align*} 
Therefore, the expectation of $z_{k,g}$ is derived as follows:
\begin{equation*}
    \mathbb{E} [z_{k,g}=1]=\frac{\text{exp}(\xi_{k,g})}{\sum_{g=1}^{G} \text{exp}(\xi_{k,g})}
\end{equation*}
,where    $\xi_{k,g}=\sum_{r=1}^{\text{R}}\text{ncol}(\text{B}_{r})(\mathbb{E}[{\ln{\gamma_{g,r}}}])+\ln(\text{det}(\text{B}_{r}))-\mathbb{E}[\gamma_{g,r}](\mu_{k,r}^{\top}\text{B}_{r}\mu_{k,r}+\text{Tr}(\text{B}_{r}\Sigma_{k,r}))$. In addition, $\mathbb{E}[{\ln{\gamma_{g,r}}}]$ equals to $\Psi(a_{g,r})-\ln{(b_{g,r})}.$ since $\gamma_{g,r}$ follows the Gamma distribution.  

\section{\emph{Inference for Eq.~\eqref{eq:20} }}\label{app:app4}
Based on Eq.~\eqref{eq:16}, $\ln{q(\alpha)}=\mathbb{E}[ \ln{p(\textbf{Y},\textbf{X}, \textbf{Z}, \Gamma,\alpha}) ]_{q(\textbf{X})q(\textbf{Z})q(\Gamma)}+const.$ Therefore,
\begin{align*}
    \ln{q(\alpha)} &\propto \mathbb{E}[ \ln{p(\textbf{Y}|\textbf{X},\alpha})p(\alpha|a,b) ]_{q(\textbf{X})} \nonumber\\
    & = \mathbb{E}[ \ln{p(\textbf{Y}|\textbf{X},\alpha)}]_{q(\textbf{X})} +\ln{p(\alpha|a,b)} \\
    & \propto (a+\frac{KM}{2}-1)\ln{\alpha}\\
    &-\alpha(b+\frac{1}{2}\sum_{k=1}^{K}(\lVert \textbf{y}_{k}
    -\Phi\mu_{k} \rVert_{2}^{2}+\text{Tr}(\Phi \Sigma_{k}\Phi^{\top}))).
\end{align*} 
Therefore, $q(\alpha)$ follows the Gamma distribution as follows:
\begin{equation*}
    q(\alpha)\sim \Gamma(\alpha| a_{\alpha}, b_{\alpha}),
\end{equation*}
where $a_{\alpha}=(a+\frac{KM}{2})$ and $b_{\alpha}=b+\frac{1}{2}\sum_{k=1}^{K}(\lVert \textbf{y}_{k}-\Phi\mu_{k} \rVert_{2}^{2}+\text{Tr}(\Phi \Sigma_{k}\Phi^{\top}))$.

\section{\emph{Inference for Eq.~\eqref{eq:27} }}\label{app:app5}
Let $Q(\textbf{B})$ be as follows:
\begin{equation*}
    Q(\textbf{B})=\mathbb{E}[ \ln{p(\textbf{X}|\textbf{Z},\Gamma;\textbf{B})} ]_{q(\textbf{X};\textbf{B}^{OLD})q(\textbf{Z};\textbf{B}^{OLD})q(\Gamma;\textbf{B}^{OLD})}.
\end{equation*}
$Q(\textbf{B})$ can be represented as follows:
\begin{align*}
    Q(\textbf{B})  &= \mathbb{E}[ \ln{p(\textbf{X}|\textbf{Z},\Gamma; \textbf{B})} ]_{q(\textbf{X})q(\Gamma)} \nonumber\\
    \begin{split}
     &= \sum_{k=1}^{K}\sum_{g=1}^{G}\mathbb{E}[z_{k,g}[\ln(\text{det}(\gamma_{g,1}\text{B}_{1})^{\frac{1}{2}}\cdots\text{det}(\gamma_{g,\text{R}}\text{B}_{\text{R}})^{\frac{1}{2}} \\
    & \text{exp}(-\frac{1}{2}\textbf{x}_{k}^{\top} \text{bdiag}[\gamma_{g,1}\text{B}_{1},...,\gamma_{g,\text{R}}\text{B}_{\text{R}}]\textbf{x}_{k})]]_{q(\textbf{X})q(\Gamma)}
    \end{split}     \\
    \begin{split}
     & = \sum_{k=1}^{K}\sum_{g=1}^{G}\mathbb{E}[z_{k,g}[\ln(\text{det}(\gamma_{g,1}\text{B}_{1}))+\cdots+\ln(\text{det}(\gamma_{g,\text{R}}\text{B}_{\text{R}})) \\
     & -(\textbf{x}_{k}^{\top} \text{bdiag}[\gamma_{g,1}\text{B}_{1},...,\gamma_{g,\text{R}}\text{B}_{\text{R}}]\textbf{x}_{k})]]_{q(\textbf{X})q(\Gamma)}  
     \end{split} \\
     \begin{split}
     & = \sum_{k=1}^{K}\sum_{g=1}^{G}\mathbb{E}[z_{k,g}\text{ncol}(\text{B}_{1})\ln(\gamma_{g,1})+\ln(\text{det}(\text{B}_{1}))+\cdots\\
     &+\text{ncol}(\text{B}_{\text{R}})\ln(\gamma_{g,\text{R}})+\ln(\text{det}(\text{B}_{\text{R}}))\\ 
     &-(\textbf{x}_{k}^{\top} \text{bdiag}[\gamma_{g,1}\text{B}_{1},...,\gamma_{g,\text{R}}\text{B}_{\text{R}}]\textbf{x}_{k})]]_{q(\textbf{X})q(\Gamma)}  
     \end{split}    \\
     \begin{split}
     & = \sum_{k=1}^{K}\sum_{g=1}^{G}\mathbb{E}[z_{k,g}]\sum_{l=1}^{\text{R}}\text{ncol}(\text{B}_{r})(\mathbb{E}[{\ln{\gamma_{g,r}}}])+\ln(\text{det}(\text{B}_{r}))\\
     &-\mathbb{E}[\gamma_{g,r}]   
     \mathbb{E}[\textbf{x}_{k,i}^{\top}\text{B}_{i}\textbf{x}_{k,i}].
     \end{split}        
\end{align*} 
Therefore, $Q(\text{B}_{i})$ is illustrated as follows:
\begin{align}
        Q(\text{B}_{i}) & = \sum_{k=1}^{K}\sum_{g=1}^{G}\mathbb{E}[ z_{k,g} ] [\ln(\text{det}(\text{B}_{i})) \nonumber\\
        &-\mathbb{E}[\gamma_{g,l}](\mathbb{E}[\textbf{x}_{k,i}^{\top}\text{B}_{i}\textbf{x}_{k,i}])]. \nonumber
\end{align}
Taking derivative to Eq.~\eqref{eq:30} with respect to $\text{B}_{i}$ leads to
\begin{align}\label{eq:31}
    \frac{\partial Q(\text{B}_{i})}{ \partial \text{B}_{i}} &=\sum_{k=1}^{K}\sum_{g=1}^{G}\mathbb{E}[ z_{k,g} ] [(\text{B}_{i})^{-1} \nonumber\\
    &-\mathbb{E}[\gamma_{g,i}](\mu_{k,i}^{\top}\mu_{k,i} +\Sigma_{k,i} )]. 
\end{align}
$\text{B}_{i}$ is estimated by letting Eq.~\eqref{eq:31} equals to zero as follows:
\begin{equation*}
    \text{B}_{i}^{-1} =\frac{\sum_{k=1}^{K}\sum_{g=1}^{G}\mathbb{E}[ z_{k,g} ]\mathbb{E}[\gamma_{g,i}](\mu_{k,i}^{\top}\mu_{k,i} +\Sigma_{k,i} )   }{\sum_{k=1}^{K}\sum_{g=1}^{G}\mathbb{E}[ z_{k,g} ] }.
\end{equation*}

\section{\emph{Fault pattern matrix $\Phi$ in Section~\ref{s:sec5}}}\label{app:app6}
\begin{figure}[!htbp]
    \centering
    \resizebox{0.5\textwidth}{!}{\includegraphics{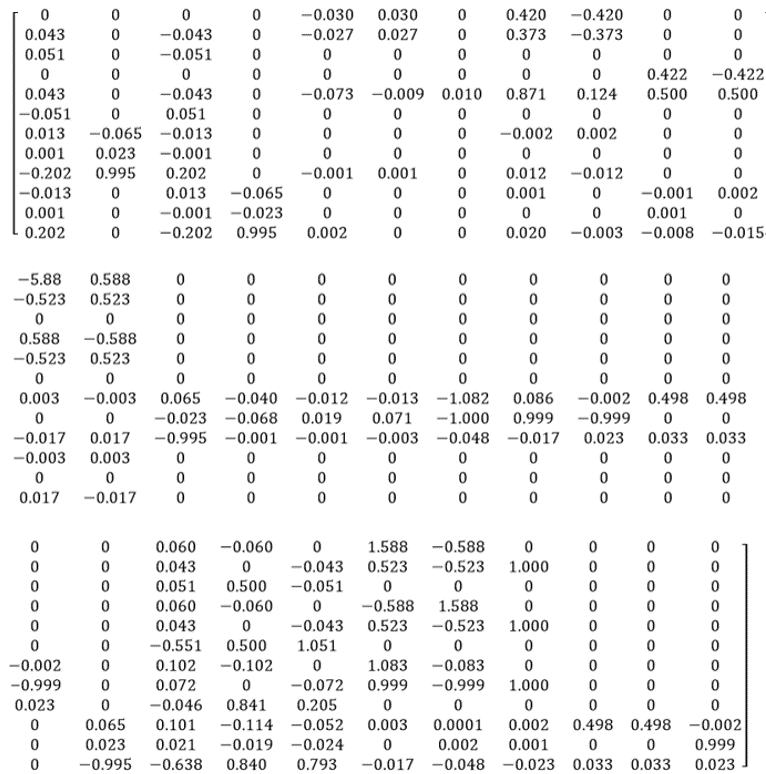}}\vspace{-0.0cm}
    \caption{Fault pattern matrix $\Phi$ in Section~\ref{s:sec5} \cite{bastani2018fault}.}
    \label{fig:fig11}
\end{figure}

Fig.~\ref{fig:fig11} shows the fault pattern matrix $\Phi$ from previous researches \cite{bastani2016compressive, huang2007stream1, kong2009variation} that used in Section~\ref{s:sec5}.



\bibliographystyle{IEEEtran}
\bibliography{TASE_IMBAGAN}

 \begin{IEEEbiography}[{\includegraphics[width=1in,height=1.25in,clip,keepaspectratio]{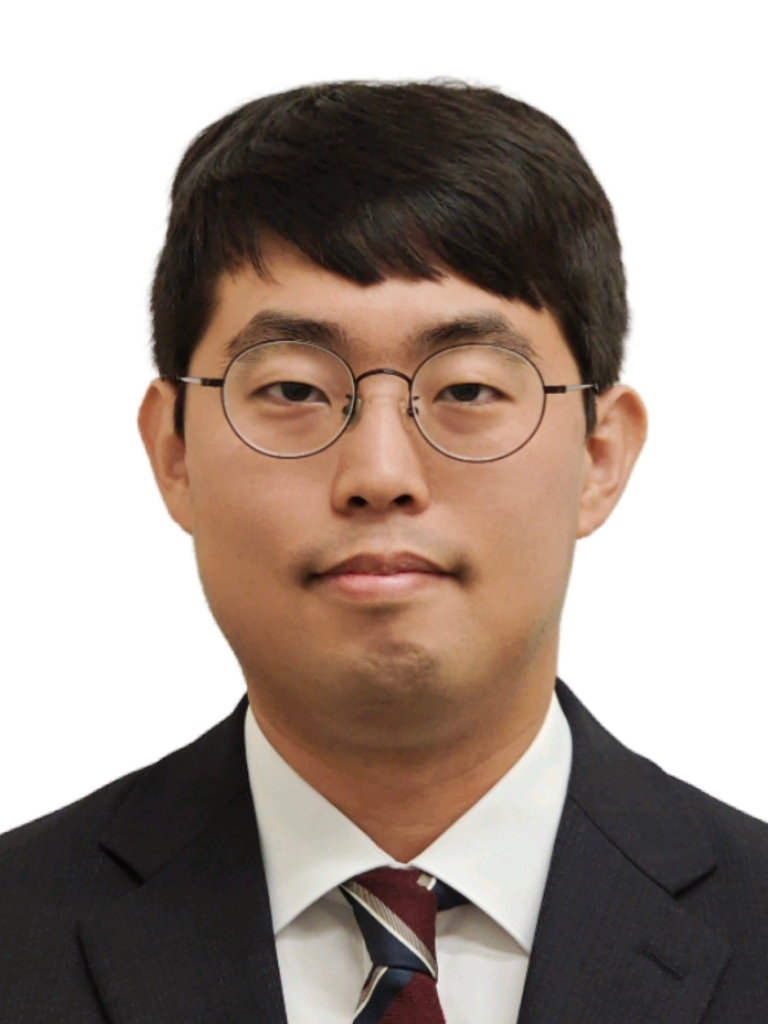}}]{Jihoon Chung } 
received his B.S. degree in Industrial Engineering from the Hanyang University, Seoul, Korea, in 2015. He obtained his M.S. degree  in Industrial and Systems Engineering at Korea Advanced Institute of Science and Technology (KAIST), Daejeon, Korea, in 2017. He received a Ph.D. degree in Industrial and Systems Engineering from Virginia Tech, Blacksburg, VA, USA. His research interests include statistical learning and data analytics in smart manufacturing.\end{IEEEbiography}
 \begin{IEEEbiography}[{\includegraphics[width=1in,height=1.25in,clip,]{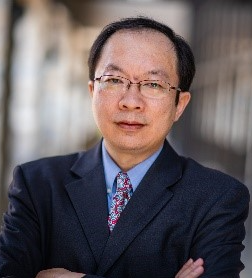}}]{Zhenyu (James) Kong}(Member, IEEE)
received the B.S. and M.S. degrees in mechanical engineering from Harbin Institute of Technology, Harbin, China, in 1993 and 1995, respectively, and the Ph.D. degree from the Department of Industrial and System Engineering, University of Wisconsin–Madison, Madison, WI, USA, in 2004. He is currently a Professor with the Grado Department of Industrial and Systems Engineering, Virginia Tech, Blacksburg, VA, USA. His research interests include sensing and analytics for smart manufacturing, and modeling, synthesis, and diagnosis for large and complex manufacturing systems.\end{IEEEbiography}
 
%






\end{document}